\documentclass[runningheads]{llncs}
\usepackage[T1]{fontenc}
\usepackage[misc]{ifsym}
\usepackage{graphicx}
\usepackage{booktabs}
\usepackage{caption}
\usepackage{subcaption}

\usepackage{amsmath,amssymb}
\usepackage{comment}

\usepackage[inline]{enumitem}
\usepackage{adjustbox}
\usepackage[algo2e,linesnumbered, ruled, vlined]{algorithm2e}
\usepackage{algpseudocode}
\usepackage{hyperref}
\usepackage{xcolor}
\usepackage{siunitx}
\newtheorem{observation}{Observation}

\usepackage{tikz}
\usetikzlibrary{matrix,fit,calc,positioning}

\graphicspath{{figures/}}

\usepackage{ifthen}
\newboolean{include-notes}
\setboolean{include-notes}{true}
\newcommand{\nb}[3]{\ifthenelse{\boolean{include-notes}}{{\colorbox{#2}{\bfseries\sffamily\scriptsize\textcolor{white}{#1}}}{\ \textcolor{#2}{\sf\small\textit{#3}}}}{}}

\newcommand{\parsDec}{\ensuremath{\theta}}  %
\newcommand{\parsEnc}{\ensuremath{\phi}}    %

\newcommand{\calD}{\ensuremath{\mathcal{D}}}
\newcommand{\calL}{\ensuremath{\mathcal{L}}}
\newcommand{\calV}{\ensuremath{\mathcal{V}}}

\newcommand{\M}{\ensuremath{\mathbf{M}}}
\newcommand{\X}{\ensuremath{\mathbf{X}}}
\newcommand{\Z}{\ensuremath{\mathbf{Z}}}

\newcommand{\missing}{\ensuremath{\perp}}

\newcommand{\vecm}{\ensuremath{\mathbf{m}}}
\newcommand{\vecx}{\ensuremath{\mathbf{x}}}
\newcommand{\x}{\vecx}

\newcommand{\vecz}{\ensuremath{\mathbf{z}}}

\newcommand{\pTrue}{\ensuremath{p^*}}

\newcommand{\elbo}{\ensuremath{\textnormal{ELBO}}}
\newcommand{\kl}{\ensuremath{\textnormal{KL}}}
\newcommand{\expectation}{\ensuremath{\mathbb{E}}}

\newcommand{\myeq}[1]{\overset{\makebox[0pt]{\tiny #1}}{=}}

\def\ci{\perp\!\!\!\perp}

\makeatletter
\def\blfootnote{\xdef\@thefnmark{}\@footnotetext}
\makeatother

\begin{document}
\title{Posterior Consistency for Missing Data in Variational Autoencoders\thanks{This work was funded in parts by the Federal Ministry of Education, Science and Research (BMBWF), Austria [Digitize! Computational Social Science in the Digital and Social Transformation].}\blfootnote{First published in Machine Learning and Knowledge Discovery in Databases: Research Track European Conference, ECML PKDD 2023, Turin, Italy, September 18–22, 2023, Proceedings, Part II, by Springer Nature. This version of the work has been extended with the addition of an Appendix, which includes proofs, the derivation of the posterior regularization, additional background information on technical topics, an extended related work section, and additional experimental results.}}
\toctitle{Posterior Consistency for Missing Data in Variational Autoencoders}

\author{Timur Sudak\inst{1} \and
Sebastian Tschiatschek\inst{1,2}(\Letter)\orcidID{0000-0002-2592-0108}} %
\authorrunning{T. Sudak and S. Tschiatschek}
\tocauthor{Timur~Sudak and Sebastian~Tschiatschek}
\institute{
University of Vienna,  Faculty of Computer Science, Vienna, Austria \and
University of Vienna, Research Network Data Science, Vienna, Austria\\
\email{sudaktimur@gmail.com,sebastian.tschiatschek@univie.ac.at}}
\maketitle              %
\begin{abstract}
   We consider the problem of learning Variational Autoencoders (VAEs), i.e., a type of deep generative model, from data with missing values.
   Such data is omnipresent in real-world applications of machine learning because complete data is often impossible or too costly to obtain.
   We particularly focus on improving a VAE's amortized posterior inference, i.e., the encoder, which in the case of missing data can be susceptible to learning inconsistent posterior distributions regarding the missingness.
   To this end, we provide a formal definition of posterior consistency and propose an approach for regularizing an encoder's posterior distribution which promotes this consistency.
   We observe that the proposed regularization suggests a different training objective than that typically considered in the literature when facing missing values.
   Furthermore, we empirically demonstrate that our regularization leads to improved performance in missing value settings in terms of reconstruction quality and downstream tasks utilizing uncertainty in the latent space.
   This improved performance can be observed for many classes of VAEs including VAEs equipped with normalizing flows.

\end{abstract}

\section{Introduction}

The availability of large amounts of data is often key to the impressive performance of nowadays machine learning (ML) models.
For instance, in computer vision, a logarithmic relationship between an ML model's performance and the amount of training data has been observed~\cite{sun2017revisiting}.
While in some cases the considered data is complete, in many relevant real-world applications, data with missing values is omnipresent. 
For instance, this holds in various applications from physical and social science~\cite{Riggi_2015,fusion_missing,social_s_missing}.
The missingness in real-world data can have different reasons, e.g., intentional or unintentional unanswered items in surveys~\cite{newman2014missing}, accidental deletion~\cite{mohan2013graphical,smith2003cost}, or high costs of exhaustive data acquisition~\cite{zhu2005cost}. 

Many approaches for the statistical analysis of data with missing values have been developed~\cite{rubin2019}. 
Some of them first complete the data (impute missing values) while others directly perform statistical analysis based on the incomplete data.
Next to classical statistical approaches, deep learning-based approaches have recently gained importance. 
Classical statistical approaches include for example mean imputation, regression-based approaches like multiple imputation by chained equations (MICE)~\cite{mice}, and MissForest~\cite{missforest}.
In the context of deep learning, recent approaches focused largely on deep generative models, in particular, generative adversarial networks (GANs) and variational autoencoders (VAEs).
GAN-based approaches include for instance generative adversarial imputation nets (GAIN)~\cite{gain} and MisGAN~\cite{misgan}.
Approaches for handling missing data in VAEs include zero-imputation VAEs (VAE-ZI)~\cite{nazabal2020handling}, partial VAEs (PVAE)~\cite{ma2019eddi}, and the missing data importance-weighted autoencoder (MIWAE)~\cite{miwae}.

In this paper, we consider and extend VAEs for dealing with data with missing values.  
In particular, we focus on an \emph{inconsistency issue} of the approximate inference network's posterior that can arise when VAEs are trained on incomplete data: the approximate inference network can be prone to produce inconsistent posterior distributions for different missingness patterns in the input data.
To overcome this issue, we identify conditions for guaranteeing \emph{posterior consistency} and propose a regularizer building on these conditions which can be integrated into the training of VAEs.
Our approach is orthogonal (and complementary) to existing approaches for VAEs for data with missing values like~\cite{nazabal2020handling,ma2019eddi,miwae}---while these approaches modify the inputs to the VAE, the structure of the approximate inference network, or the lower bound on the likelihood to be optimized, they do not add explicit regularization on the approximate posterior distributions regarding the missingness in the input. 
Our approach also differs from other works that have considered forms of posterior consistency~\cite{abstract_liu,sinha2021consistency,zhu_nlp_reg_2022}---these works mainly build on the idea that transformed inputs should map to similar posterior distributions which is not necessarily implied by our approach building on first principles.
We empirically demonstrate that our proposed regularization leads to improved imputation performance and improved performance in downstream applications in missing data settings.    

Our contributions are:
\begin{itemize}
    \item We propose a notion of \emph{posterior consistency} and show its importance for VAEs for data with missing values.
    
    \item We propose a regularizer for promoting the posterior consistency of VAEs. 
    
    \item We empirically demonstrate the superior performance of VAEs trained with our proposed regularization in comparison to many natural baselines. 

    \item The source code for reproducing our experiments is available on github.\footnote{\href{https://github.com/stschia/VAE-posterior-consistency.git}{https://github.com/stschia/VAE-posterior-consistency.git}}
\end{itemize}

Our paper is organized as follows:
In Section~\ref{sec:background} we introduce the relevant background and notation before describing the problem of posterior inconsistency in Section~\ref{sec:problem}.
We describe our approach in Section~\ref{sec:methodology}, followed by a discussion of related work in Section~\ref{sec:related}.
In Sections~\ref{sec:experimental-setup} and \ref{sec:experiments} we empirically evaluate our proposed approach.
We conclude our paper in Section~\ref{sec:conclusions}.

\section{Background}
\label{sec:background}

In this section, we introduce our notation and the necessary background about VAEs and the considered types of missing data. 

\paragraph{Notation.} 
We use uppercase letters to denote random variables (RVs), e.g., $X$, and lowercase letters to denote instantiations of RVs, e.g., $x$.
We use bold uppercase letters to denote vectors of RVs, e.g., $\X$, and bold lowercase letters to denote their respective instantiations, e.g., $\vecx$.
Furthermore, we use $P$ and $Q$ to denote subsets of some fixed ground set $\mathcal{V} = \{1, \ldots, d\}$, i.e., $P,Q \subseteq \mathcal{V}$.
Assuming $\X = [X_1, \ldots, X_d]^T$, we use $\X_Q$ to denote the vector of random variables $[X_{i_1}, \ldots, X_{i_k}]^T$, where $|Q| = k$ and $Q = \{i_1, \ldots, i_k\}$.
We denote the $i$th component of $\vecx$ by $x_i$ and, similarly to before, denote by $\vecx_Q$ the vector $[x_{i_1}, \ldots, x_{i_k}]^T$.

\paragraph{Data.}
Assume $n$ i.i.d.\ samples of dimension $d$ from an unknown data distribution $\pTrue$, i.e., $\tilde{\vecx}^1, \ldots, \tilde{\vecx}^n \sim p^*$.
For those samples, only a subset of the dimensions (features) is available to us, i.e., for a sample $\tilde{\x}^i$ there exists $Q^i \subseteq \mathcal{V}$ and only $\smash{\tilde{\vecx}^i_{Q^i}}$ is available to us.
The set of present features $Q^i$ can depend on $\tilde{\vecx}^i$ in different ways according to an unknown missingness mechanism (details are at the end of this section).
We collect the partial data into the data set $\mathcal{D} = \{ \tilde{\vecx}^1_{Q^1}, \ldots, \tilde{\vecx}^n_{Q^n}\}$.
Equivalently we can consider $\mathcal{D} = \{\vecx^1, \ldots, \vecx^n\}$, where $\vecx^i$ are of dimension $d$ and missing dimensions contain the special symbol $\missing$.
To simplify notation, we will assume that $\missing$ always assumes the implied dimensionality, e.g., a statement like $\vecx_Q = \missing$ implies that $\missing$ is a vector of size $|Q|$ in which each dimension is $\missing$.

\paragraph{Variational autoencoders.}
A common approach to modeling complicated distributions $\pTrue$ are VAEs, a type of deep generative latent variable models~\cite{vae13}.
In vanilla VAEs, one assumes the data to be generated as follows: %
\begin{equation*}
    \vecz \sim p(\Z), \quad
    \tilde{\vecx} \sim p_{\parsDec}(\tilde{\X} | \vecz),
\end{equation*}
where $\vecz$ are latent variables following a prior distribution $p(\Z)$, commonly assumed to be the normal distribution of dimension $k$ with diagonal covariance matrix, and $p_{\parsDec}(\tilde{\X} | \vecz) = \mathcal{N}(\mu_{\parsDec}(\vecz), \textnormal{diag}(\sigma_{\parsDec}^2(\vecz) ))$ is a normal distribution of dimension $d$, where ${\parsDec}$ denotes the parameters of neural networks parameterizing its mean as $\mu_{\parsDec}\colon \mathbb{R}^k \rightarrow \mathbb{R}^d$ and its standard deviation as  $\sigma_{\parsDec}\colon \mathbb{R}^k \rightarrow \mathbb{R}^d$. 
The generative model induces a distribution over $\tilde{\X}$ through marginalization over $\Z$, i.e., $p_\parsDec(\tilde{\X}) = \int_{\vecz} p_\parsDec(\tilde{\X}|\Z=\vecz)p(\Z=\vecz) \, \textnormal{d}\vecz$.
We often drop the subscript $\parsDec$ for brevity.

VAEs are typically fit to data $\calD$ by maximizing a lower bound on the marginal log-likelihood $\mathcal{L}(\mathcal{D}) = \sum_i \log p_\parsDec(\tilde{\vecx}^i)$, the so-called \emph{evidence lower bound} (ELBO):
\begin{align*}
    \calL(\calD) \geq \sum_{i=1}^n\! \big[ \expectation_{\vecz \sim q_{ \parsEnc}(\Z | \tilde{\vecx}^i)}[\log p_{\parsDec}(\tilde{\vecx}^i | \vecz)] - \kl(q_{\parsEnc}(\Z | \tilde{\vecx}^i) \| p(\Z)) \big],
\end{align*}
where $q_{\parsEnc}(\Z | \tilde{\vecx}^i)$ is an approximation to the true posterior distribution $p_\parsDec(\Z | \tilde{\vecx}^i)$, implemented by a neural network with parameters $\parsEnc$, the so-called inference network or encoder.
Abusing notation, we will in the following often write $q_{\parsEnc}(\tilde{\vecx}^i)$ instead of $q_{\parsEnc}(\Z | \tilde{\vecx}^i)$.
In vanilla VAEs, $q_{\parsEnc}(\tilde{\vecx}^i)$ is a normal distribution of dimension $k$ with a diagonal covariance matrix whose parameters are computed by the encoder.
The type of distribution of $q_{\parsEnc}(\tilde{\vecx}^i)$ is also referred to as the variational family, and the encoder is said to perform amortized inference of the posterior as the parameters of the variational family's distribution are predicted by a neural network and not optimized on a sample by sample basis~\cite{zhang2018advances}.
Typically, the ELBO is not computed exactly but a stochastic approximation of the ELBO is considered. 
For effective learning, the reparametrization trick is leveraged~\cite{vae13}.
The difference between the marginal log-likelihood and the ELBO depends, despite other things, on the used variational family, i.e., the choice of distribution for $q_{\parsEnc}(\tilde{\vecx})$, and the quality of the amortization~\cite{cremer2018inference}.

\paragraph{Missing data.}
Missing data is a common phenomenon in many real-world settings. 
In the literature, different types of missingness are distinguished depending on how the missing variables are determined~\cite{rubin_inference,little2019statistical}.
To introduce those, let $Q$ be the set of available features of sample $\vecx$ and denote by $\tilde{\vecx}$ the corresponding complete sample.
The following missingness mechanisms are typically considered:
\begin{itemize}
    \item \emph{Missing completely at random (MCAR)}: 
    The data is MCAR if $Q$ is independent of $\tilde{\vecx}$, i.e., $Q \ci \tilde{\vecx}$, where $\ci$ denotes conditional independence.
    
    \item \emph{Missing at random (MAR)}: The data is MAR if its missingness can be explained solely by the observed variables, i.e., $\tilde{\vecx}_{\calV - Q} \ci Q \mid \tilde{\vecx}_{Q}$. %
    
    \item \emph{Missing not at random (MNAR)}: Any missingness which is not MCAR or MAR is  MNAR. 
\end{itemize}

\paragraph{Variational autoencoders for missing data.} Models can be fit to data with missingness of MCAR or MAR type by maximizing the likelihood of the data (the missingness is ignorable \cite{rubin_inference}).
This has been exploited for training VAEs from incomplete data by extending the ELBO (e.g., \cite{ma2019eddi,nazabal2020handling}) such that 
\begin{align*}
    \calL(\calD) \! \geq \! \sum_{i=1}^n \! \big[ \expectation_{\vecz \sim q_{\parsEnc}(\vecx^i_{Q_i})}[\log p_{\parsDec}(\vecx^i_{Q^i} | \vecz)] - \kl(q_{\parsEnc}(\vecx^i_{Q^i}) \| p(\Z)) \big].
    \label{eq:elbo-partial}
\end{align*}
In the case of the zero-imputation VAE (VAE-ZI), the encoder $q_{\parsEnc}(\cdot)$ is provided with the partially-observed input $\vecx^i$ with missing features replaced by zeros. Zero imputation of the input is also used for the MIWAE but in contrast to the vanilla VAE a tighter lower bound to the data's likelihood based on $M$ importance weighted samples $\mathbf{z}_1,\ldots, \mathbf{z}_M \sim q_{\phi}(\vecx^i_{Q^i})$ is considered:
\begin{align*} \calL^M(\calD) \geq    \sum_{i=1}^n \Bigg[\mathbb{E}_{\mathbf{z}_{1}, \ldots, \mathbf{z}_{M} \sim q_\phi(\vecx^i_{Q^i})}\Big[\log \frac{1}{M} \sum_{k=1}^M \frac{p_{\boldsymbol{\theta}}(\vecx^i_{Q^i} | \mathbf{z}_{k}) p(\mathbf{z}_{k})}{q_\phi(\vecx^i_{Q^i})}\Big] \Bigg].
\end{align*}
For PVAEs, a permutation-invariant set function is used as the encoder, i.e., $q_{\parsEnc}(\vecx^i_{Q^i}) = g(h(\mathbf{s}_1), h(\mathbf{s}_2), \ldots, h(\mathbf{s}_{|Q^i|}))$, where each $\mathbf{s}_j$ is the concatenation or multiplication of a learned embedding $\mathbf{e}_j$ and the corresponding observed feature $x^i_{j}$, and the permutation-invariant function $g(\cdot)$ is the summation of outputs from neural networks $h(\cdot)$, potentially followed by further neural network layers.

\section{Posterior Inconsistency}
\label{sec:problem}

During training, the approximate posterior $q_{\parsEnc}(\vecx)$ of a VAE is fit to the generative model specified through $p(\vecz)$ and $p_{\parsDec}(\vecx | \vecz)$ via maximization of the ELBO.
Typically $q_{\parsEnc}(\vecx)$ does not perfectly match the true posterior distribution $p_{\parsDec}(\Z | \vecx)$ because the  encoder's variational family is not expressive enough (``approximation gap'') and because the amortization results in suboptimal predictions for the parameters of the variational distributions (``amortization gap'')~\cite{cremer2018inference}.
In this paper, we focus on special aspects of these gaps which occur when working with missing data and which can significantly decrease performance in downstream tasks.

We refer to the problem under consideration as \emph{posterior inconsistency} and broadly use this term to denote inconsistencies in the approximate posterior distribution when applying VAEs in cases of incomplete data.
In such cases, a VAE's approximate posterior is computed from a partial sample $\vecx_Q$, for some $Q \subseteq \mathcal{V}$~\cite{miwae,ma2019eddi,nazabal2020handling}.
Importantly, there is a strong dependency between the posteriors $p_\parsDec(\vecz | \vecx_Q)$ and $p_\parsDec(\vecz | \vecx_P)$ for $P \subseteq Q$ that should be reflected in the approximate posterior but often this is not the case.

\begin{figure*}[!tb]
  \centering
  \setlength{\belowcaptionskip}{-0.2\baselineskip}
  \begin{subfigure}[t]{0.14\textwidth}
    \includegraphics[width=1.2\textwidth]{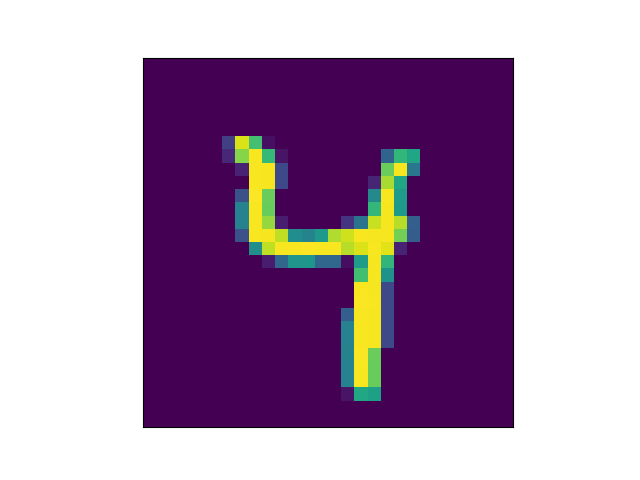}
    \subcaption{Original}
    \label{fig:mot:original}
  \end{subfigure}
  \begin{subfigure}[t]{0.14\textwidth}
    \includegraphics[width=1.2\textwidth]{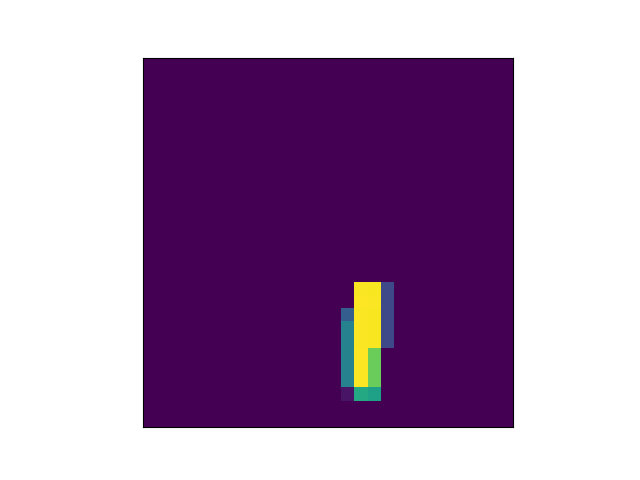}
    \subcaption{Masked}
    \label{fig:mot:masked}
  \end{subfigure}
  \begin{subfigure}[t]{0.23\textwidth}
    \centering
    \includegraphics[width=0.8\textwidth]{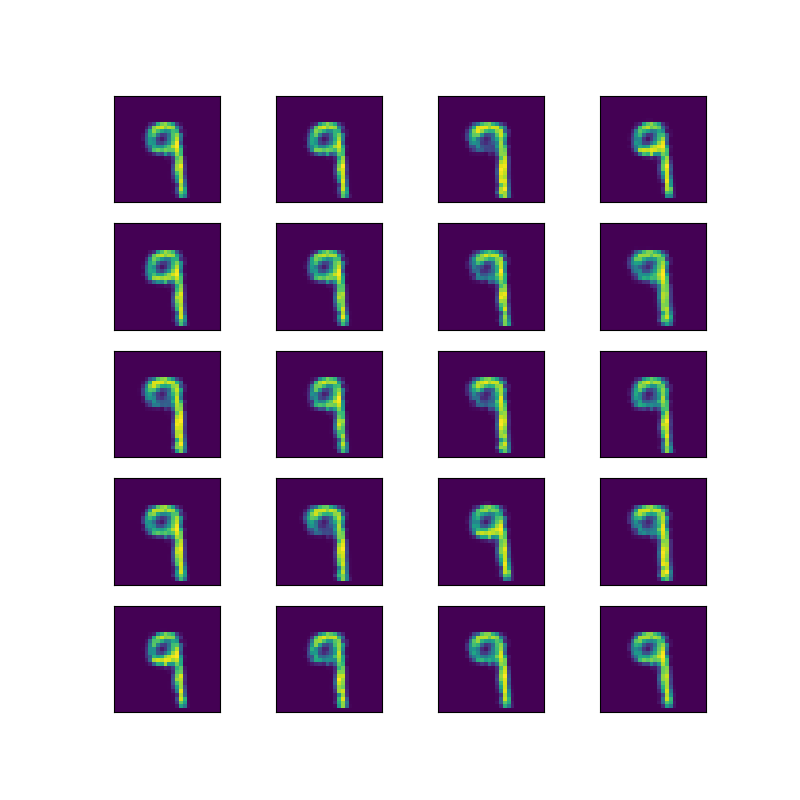}
    \subcaption{AM-VAE-PNP}
    \label{fig:mot:AM-VAE-PNP}
  \end{subfigure}%
  \begin{subfigure}[t]{0.23\textwidth}
    \centering
    \includegraphics[width=0.8\textwidth]{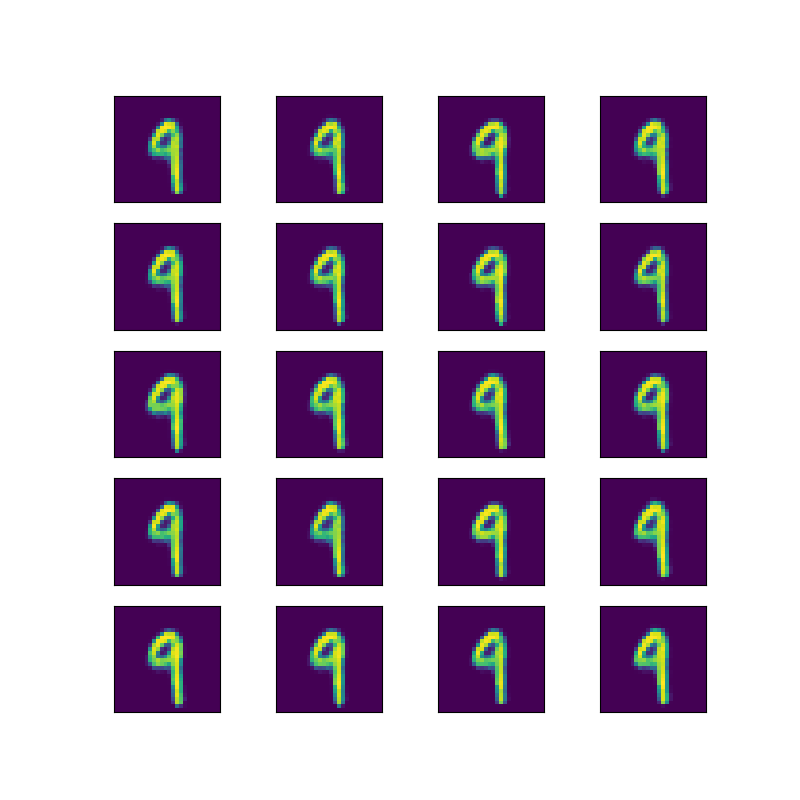}
    \subcaption{VAE-PNP \phantom{asdfs}}
    \label{fig:mot:VAE-PNP}
  \end{subfigure}%
  \begin{subfigure}[t]{0.23\textwidth}
    \centering
    \includegraphics[width=0.8\textwidth]{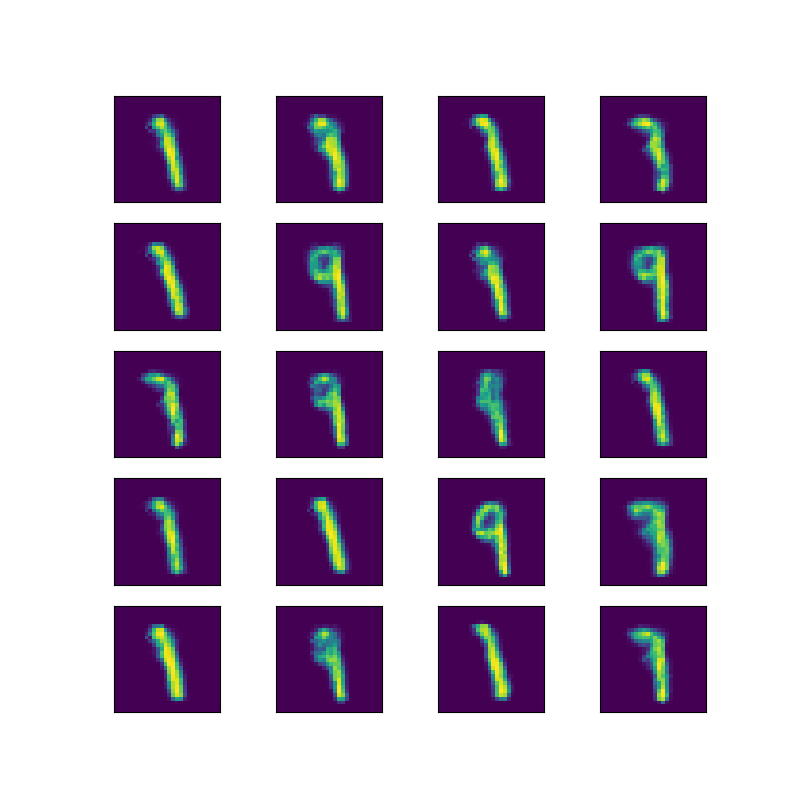}
    \subcaption{\mbox{REG-VAE-PNP} (ours)}
    \label{fig:mot:REG-VAE-PNP}
  \end{subfigure}
  \caption{Imputation of an image for which the upper part is missing. \textit{(\subref{fig:mot:original})} Original image. 
  \textit{(\subref{fig:mot:masked})} Image with masked (missing) upper part. 
  \textit{(\subref{fig:mot:AM-VAE-PNP})} \textit{(\subref{fig:mot:VAE-PNP})} \textit{(\subref{fig:mot:REG-VAE-PNP})} Images imputed by AM-VAE-PNP, VAE-PNP, and REG-VAE-PNP (ours) for the masked input image, respectively. The model with our proposed regularization produces different plausible completions while models without our regularization produce almost deterministic outputs. See Section~\ref{sec:experimental-setup} for details regarding the models. 
  }
  \label{fig:mnist_motivation2}
\end{figure*}

The consequences of such inconsistency can be observed in the task of image inpainting, cf.\ Figure~\ref{fig:mnist_motivation2}. 
As shown in Figures~\ref{fig:mot:AM-VAE-PNP} and~\ref{fig:mot:VAE-PNP}, models that do not accurately reflect the relationship between $p_\parsDec(\vecz | \vecx_P)$ and $p_\parsDec(\vecz | \vecx_Q)$ can suffer from posterior inconsistency, which in turn can cause overconfident image imputation. 
By contrast, see Figure~\ref{fig:mot:REG-VAE-PNP}, models trained with regularization of the relationship of the approximate posteriors for different subsets of missing features can provide more robust imputations, i.e., better account for different plausible imputations.

This posterior inconsistency can lead to reduced performance in downstream tasks if they rely on computations in the latent space.
For example, previous work has considered the computation of information rewards based on the approximate posteriors in the latent space in order to select which feature to acquire in active feature selection scenarios~\cite{ma2019eddi}.
Clearly, the accuracy of such a computation strongly depends on the quality of the approximate posterior distribution, and a bad approximate posterior distribution can result in incorrect rewards for the different variables that could be selected, cf.\ our experiments in Section~\ref{sec:experiments}.

In the next section, we formally define \emph{posterior consistency} and propose an effective regularizer for improving the consistency of the approximate posterior under missingness which helps to alleviate the problems sketched above.

\section{Methodology}
\label{sec:methodology}

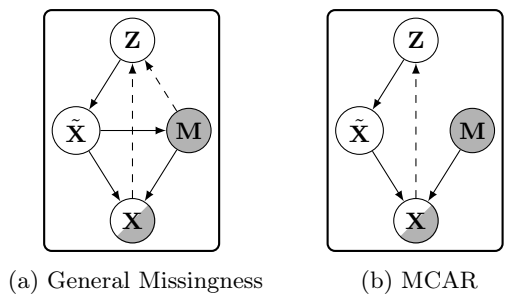
\begin{figure}[!htbp]
    \centering
     \begin{subfigure}[b]{0.3\textwidth}
         \centering
         \colorlet{myblack}{black!30}

\begin{tikzpicture}

\tikzstyle{surround} = [thick,draw=black,rounded corners=1mm]
\tikzstyle{scalarnode} = [circle, draw, 
    text width=1.2em, text badly centered, inner sep=1.5pt]
\tikzstyle{scalarnodenoline} = [  fill=white!11, 
    text width=1.2em, text badly centered, inner sep=2.5pt]
\tikzstyle{arrowline} = [draw,color=black, -latex]
\tikzstyle{dashedarrowcurve} = [draw,color=black, dashed, -latex]
\tikzstyle{dashedarrowline} = [draw,color=black, dashed,  -latex]

    \node [scalarnode] at (0, 0) (Z) {$\Z$};
    \node [scalarnode] at (-0.75, -1.2) (Xtilde) {$\tilde{\X}$};
    \node [scalarnode, fill=myblack] at (0.75, -1.2) (M) {$\M$};

    \draw[fill=myblack,draw=none] (0.0,-2.4) -- ++(45:3.05mm) arc (45:-135:3.05mm) -- cycle ;
    \node [scalarnode] at (0.0, -2.4) (X) {$\X$};

    \node[surround, inner sep = .1cm] (f_N) [fit = (Z)(X)(Xtilde)(M) ] {};
    \path[dashedarrowcurve]  (X) to (Z);
    \path[dashedarrowcurve]  (M) to (Z);

    \path [arrowline] (Z) to (Xtilde);
    \path [arrowline] (Xtilde) to (M);
    \path [arrowline] (Xtilde) to (X);
    \path [arrowline] (M) to (X);
\end{tikzpicture}
         \caption{General Missingness}
         \label{fig:y equals x}
     \end{subfigure}
     \begin{subfigure}[b]{0.3\textwidth}
         \centering
         \begin{tikzpicture}

\colorlet{myblack}{black!30}

\tikzstyle{surround} = [thick,draw=black,rounded corners=1mm]
\tikzstyle{scalarnode} = [circle, draw, fill=white!11,  
    text width=1.2em, text badly centered, inner sep=1.5pt]
\tikzstyle{scalarnodenoline} = [  fill=white!11, 
    text width=1.2em, text badly centered, inner sep=1.5pt]
\tikzstyle{arrowline} = [draw,color=black, -latex]
\tikzstyle{dashedarrowcurve} = [draw,color=black, dashed, out=90,in=-90, -latex]
\tikzstyle{dashedarrowline} = [draw,color=black, dashed,  -latex]

    \node [scalarnode] at (0, 0) (Z) {$\Z$};
    \node [scalarnode] at (-0.75, -1.2) (Xtilde) {$\tilde{\X}$};
    \node [scalarnode, fill=myblack] at (0.75, -1.2) (M) {$\M$};

    \draw[fill=myblack,draw=none] (0.0,-2.4) -- ++(45:3.05mm) arc (45:-135:3.05mm) -- cycle ;
    \node [scalarnode, fill=none] at (0.0, -2.4) (X) {$\X$};

    \node[surround, inner sep = .1cm] (f_N) [fit = (Z)(X)(Xtilde)(M) ] {};
    \path[dashedarrowcurve]  (X) to (Z);

    \path [arrowline] (Z) to (Xtilde);
    \path [arrowline] (Xtilde) to (X);
    \path [arrowline] (M) to (X);
\end{tikzpicture}
         \caption{MCAR}
         \label{fig:three sin x}
     \end{subfigure}
    \caption{Graphical models representing the generative model of the data and the inference models (dashed lines) used in the VAEs.
    The model consists of latent variables $\Z$, the complete sample $\tilde{\X}$, the missingness pattern $\M$, and the partially observed data $\X$.
    In the case of general missingness, the missingness pattern can contain valuable information about the latent variables $\Z$.
    Half-shaded circles are used to emphasize that $\X$ contains missing values according to $\M$.}
    \label{fig:graphical-models}
\end{figure}

Our approach to improving the performance of VAEs in the face of missing data builds on the relationship of the posterior distributions for different sets of available features.
In particular, let $P, Q \subseteq \mathcal{V}$ and $P \subseteq Q$ denote two subsets of the available features, and let $\bar{P} = Q \setminus P$, i.e., $P$ contains only a subset of the features of $Q$ and $\bar{P}$ is the set of features in $Q$ which are not in $P$.
We are interested in understanding the relationship of $p_\theta(\vecz | \vecx_P)$ and $p_\theta(\vecz | \vecx_Q)$.
To this end, we assume the generative model according to Figure~\ref{fig:graphical-models} for the data and the missingness, i.e.,
\begin{equation*}
    \vecz \sim p(\Z),\quad
    \tilde{\vecx} \sim p_{\parsDec}(\tilde{\X} | \vecz),\quad
    \vecm \sim p_{\parsDec}(\mathbf{M} | \tilde{\vecx}), \quad
    \vecx \sim p_{\parsDec}(\X | \tilde{\vecx}, \vecm).
\end{equation*}
For MCAR data, $\vecm \sim p_{\parsDec}(\mathbf{M} | \tilde{\vecx})$ reduces to $\vecm \sim p_{\parsDec}(\mathbf{M})$.
The observed $\vecx$ is a deterministic function of $\tilde{\vecx}$ and $\vecm$, i.e., $p_{\parsDec}(\X | \tilde{\vecx}, \vecm)$ has all its probability mass on the single $\vecx$ corresponding to $\tilde{\vecx}$ where missing values according to $\vecm$ are replaced by $\missing$.
We additionally make the common assumption that given $\vecz$ the dimensions of $\tilde{\vecx}$ are generated independently.

For this generative model, we can make the following observation (see Appendix~\ref{sec:posterior-consistency-observation} for details) regarding the relationship of the posteriors for different sets of available features:
\begin{observation} \label{obs:posterior}
For missing data we have
\begin{equation*}
  p(\vecz | \vecx_Q,  \vecx_{\calV-Q}=\perp)=p(\vecz | \vecx_P, \vecx_{\calV-Q}=\perp )  \Big[ p(\vecx_{\bar{P}} | \vecz, \vecx_P, \vecx_{\calV-Q}=\perp )  \frac{p(\vecx_P, \vecx_{\calV-Q}=\perp )}{p(\vecx_Q,  \vecx_{\calV-Q}=\perp)} \Big].
\end{equation*}
For MCAR data this simplifies to
\begin{equation}
\label{eq:mcar_kl_equation}
  p(\vecz | \vecx_Q) = p(\vecz| \vecx_P) \Big[ \frac{p(\vecx_{P})}{p(\vecx_Q)} p(\vecx_{\bar{P}} | \vecz) \Big] .
\end{equation}
\end{observation}

For brevity of notation, we continue the exposition of our approach using the notation for the MCAR case but it would be analogous for MAR and MNAR.
Based on the above observation, we define the notion of \emph{posterior consistency}.\\
\begin{definition}[Posterior consistency]
A family of conditional distributions $\{ \psi(\Z | \vecx_Q) | \forall \vecx , Q \subseteq \calV\}$ is \emph{posterior consistent} with the generative model $p_\theta$ and possible missingness
if $\forall \vecx_Q \sim p_\theta(\X) \; \forall P \subseteq Q \subseteq \calV$ with $p_\theta(\vecx_P) > 0$,
 \begin{align}
    \label{eq:full_kl}
    \kl \Big( \psi(\Z | \vecx_Q) \| \frac{p(\vecx_P)}{p(\vecx_Q)} \cdot p(\vecx_{\bar{P}} | \Z) p_\theta(\Z | \vecx_P)\Big)
  = 0.
  \end{align}%
\end{definition}
This definition relates the posteriors $p_\theta(\vecz | \vecx_P)$ and  $p_\theta(\vecz | \vecx_Q)$ for different nested subsets of features $P$ and $Q$.
Clearly, the true posterior is posterior-consistent. 
Note that if \eqref{eq:full_kl} is zero, then 
\begin{align*}
 \log p(\vecx_Q)=&\log p(\vecx_P)+\mathbb{E}_{\vecz \sim \psi(\Z | \vecx_Q)} \log p(\vecx_{\bar{P}} | \vecz)- \kl(\psi(\Z | \vecx_Q) \| p_\theta(\Z | \vecx_P)).
\end{align*}
Importantly, for the special case $\psi(\Z | \vecx) = p_\parsDec(\Z | \vecx)$,
\begin{align}
 \log p(\vecx_Q)=&\log p(\vecx_P)\!+\!\mathbb{E}_{\vecz \sim p_{\theta}(\Z | \vecx_Q)} \log p(\vecx_{\bar{P}} | \vecz)- \kl(p_\theta(\Z | \vecx_Q) \| p_\theta(\Z | \vecx_P)).
  \label{clear_form}
\end{align}
Thus, given posterior consistency, the marginal log-likelihoods of $\vecx_Q$ and  $\vecx_P$ are closely related.
Existing approaches for dealing with partial data in VAEs approximate the marginal log-likelihood of $\vecx_Q$ or $\vecx_P$ without considering their relation according to \eqref{clear_form}. For example,  in \cite{nazabal2020handling} only $\vecx_Q$ was considered, whereas in  \cite{ma2019eddi} $\vecx_P$ was introduced during training but no connection between $\vecx_Q$ and $\vecx_P$ is explicitly considered.

As mentioned, posterior consistency obviously holds for the true posterior $p_\theta(\Z | \vecx_Q)$.
In VAEs, this posterior is approximated by $q_\phi(\Z | \vecx_Q)$. 
Thus it would be sensible to require also the approximate posterior to satisfy
\begin{equation*}
  \kl \Big(q_\phi(\Z | \vecx_Q) \| \frac{p(\vecx_P)}{p(\vecx_Q)} \cdot p(\vecx_{\bar{P}} | \Z) p_\theta(\Z | \vecx_P)\Big)
  = 0.
  \label{eq:approximteConsistentcy}
\end{equation*}
Unfortunately, evaluating this $\kl$-divergence is infeasible as it would require computing the true model's posterior. 
Therefore, we use the following proxy:
\begin{equation*}
	\kl \Big(q_\phi(\Z | \vecx_Q) \| \frac{p_{\mathcal{\theta}}(\vecx_P)}{p_{\mathcal{\theta}}(\vecx_Q)} \cdot p_{\theta}(\vecx_{\bar{P}} | \Z) q_\phi(\Z | \vecx_P)\Big)
\end{equation*}
Note that this form of posterior consistency might not be achievable  because of the limited expressiveness of the approximate posteriors when using an insufficiently expressive variational family.
But even in such a case, we might hope for a better alignment of the approximate posterior for different missingness patterns resulting in better empirical performance, cf. experiments in Section~\ref{sec:experiments}.

We include the requirement for posterior consistency in the process of training the encoder and decoder of the VAE by maximizing the following objective (expressed for a single sample $\vecx^i$ for brevity): 
\begin{align} \label{eq:elboext-main}
 \mathcal{L}_{\lambda, \theta, \phi} =  & \mathcal{L}_{\theta, \phi}^{\elbo}(\vecx^i_{Q^i}) - \lambda \Big[ \kl (q_\phi (\Z | \vecx^i_{Q^i} ) \| q_\phi (\Z | \vecx^i_{P^i} ) ) \\
  & - \mathbb{E}_{\vecz \sim q_{\phi} (\Z \mid \vecx^i_{Q^i})} \log p_\theta (\vecx_{\bar{P}^i}^i | \Z )  \nonumber  - \log \frac{p_{\mathcal{\theta}}(\vecx^i_{P^i})}{p_{\mathcal{\theta}}(\vecx^i_{Q^i})}  \Big] \nonumber 
\end{align}
where $\lambda$ is a hyper-parameter allowing a trade-off between ELBO maximization and posterior consistency and $P^i$ is a random subset of $Q^i$ (details below).
The ELBO term $\mathcal{L}_{\theta, \phi}^{\elbo}(\vecx^i_{Q^i})$ for the partial sample $\vecx^i_{Q^i}$ is given as 
\begin{align*}
\mathcal{L}_{\theta, \phi}^{\elbo}(\vecx^i_{Q^i})  = \expectation_{\vecz \sim q_{\parsEnc}(\vecx^i_{Q^i})}[\log p_{\parsDec}(\vecx^i_{Q^i} | \vecz)] \!-\! \kl(q_{\parsEnc}(\vecx^i_{Q^i}) \| p(\Z)).
\end{align*}
Furthermore, we approximate the expression \smash{$\log \frac{p_{\mathcal{\theta}}(\vecx^i_{P^i})}{p_{\mathcal{\theta}}(\vecx^i_{Q^i})}$} in~\eqref{eq:elboext-main} by
$\mathcal{L}_{\theta, \phi}^{\elbo}(\vecx^i_{P^i})- \mathcal{L}_{\theta, \phi}^{\elbo}(\vecx^i_{Q^i}),$
where the quality of this approximation will depend on the class of used VAE models and the parameters $\parsDec, \parsEnc$.
Models with more expressive posteriors, e.g., normalizing flows~\cite{rezende_flow_15}, typically have a smaller inference gap~\cite{cremer2018inference} and are likely to lead to a better approximation.
For a more detailed derivation of the above objective please refer to Appendix~\ref{sec:reg_derivation}.

\paragraph{Training VAEs for posterior consistency.}
Our approach for optimizing~\eqref{eq:elboext-main} is presented in Algorithm~\ref{alg:cap}.
Note that artificial missingness is added to create the sets of features $P^j$ from $Q^j$ (line~\ref{alg:missingness}).
In the simplest case, features are removed randomly with some fixed probability but other schemes, e.g., respecting a learned or the true missingness mechanisms, are possible.
New missingness patterns are created randomly in each iteration.
Even for the simple case of removing features with a fixed probability, we observed empirically that our regularization not only improves performance for MCAR data or MAR data, where the missingness mechanism can in principle be ignored \cite{rubin_inference} but also for some MNAR scenarios, cf.\ our Experiments in Section~\ref{sec:experiments}.

\algnewcommand\algorithmicinput{\textbf{Input:}}
\algnewcommand\INPUT{\item[\algorithmicinput]}

\newcommand\mycommfont[1]{\footnotesize\ttfamily\textcolor{blue}{#1}}
\SetCommentSty{mycommfont}

{
\setlength{\abovedisplayskip}{2pt}
\setlength{\belowdisplayskip}{-6pt}

\begin{algorithm2e}[!h]
\SetAlgoLined
\DontPrintSemicolon

\caption{Training algorithm for regularized VAE}\label{alg:cap}
\KwIn{Partially observed training data $\mathcal{D}$, regularization parameter $\lambda$, and percentage of missingness $\mathcal{P}$}
Initialize $\parsDec$ and $\parsEnc$\;
\For{$t = 1,2,\ldots$}{
  \tcc{Get data and create artificial missingness}
  Obtain indices $I$ for minibatch \;
  Obtain minibatch
    $\mathcal{B}_Q = \{ \vecx^j_{Q^j} | j \in I \}$\;
  Remove features with probability $\mathcal{P}$: \hfill
  $\mathcal{B}_P = \{ \vecx^{j}_{P^{j}} | \vecx^j_{Q^j}, j \in I, \textnormal{$P^j$  random subset of $Q^j$} \}$ \label{alg:missingness}\;
  $\mathcal{B}_{\bar{P}} = \{ \vecx^{j}_{\bar{P}^{j}} | j \in I, \bar{P}^j = Q^j \setminus P^j \}$ \;
  $\mathcal{Z} = \{ \vecz^j_{Q^j} \sim  \mathcal{N}(\mu_{\parsEnc}(\vecx^j_{Q^j}), \Sigma_{\parsEnc}(\vecx^j_{Q^j})) \mid j \in I \}$ \;
  $\mathcal{Z}' = \{ \vecz^j_{P^j} \sim  \mathcal{N}(\mu_{\parsEnc}(\vecx^j_{P^j}), \Sigma_{\parsEnc}(\vecx^j_{P^j})) \mid j \in I \}$ \;
  \tcc{Compute loss as in Equation~\eqref{eq:elboext-main}}
  Compute $\mathcal{L}^{\elbo}_{\parsDec,\parsEnc}(\mathcal{B}_Q)=\frac{1}{|\mathcal{B}_Q|} \sum_{j\in I} [\log p_{\parsDec}(\vecx^{j}_{Q^{j}}, \vecz^{j}_{Q^{j}} )-\log q_{\parsEnc}(\vecz^{j}_{Q^{j}} \mid \vecx^{j}_{Q^{j}})]$\;
  Compute $\mathcal{L}^{\elbo}_{\parsDec,\parsEnc}(\mathcal{B}_P) =
      \frac{1}{|\mathcal{B}_P|} \sum_{j\in I}[\log p_{\parsDec}(\vecx^{j}_{P^{j}}, \vecz^{j}_{P^{j}})-\log q_{\parsEnc}(\vecz^{j}_{P^{j}} \mid \vecx^{j}_{P^{j}})]$\;
  Compute log-likelihood $\ell(\mathcal{B}_{\bar{P}} ; \theta) =
    \frac{1}{|\mathcal{B}_{\bar{P}}|} \sum_{j \in I} \log p_{\parsDec}(\vecx_{\bar{P}^j}^j | \vecz_{Q^j}^j )$\;
  Compute KL-divergence \hfill%
  $
  \qquad r = \sum_{j \in I} \kl(\mathcal{N}(\Z|\mu_{\parsEnc}(\vecx^j_{Q^j}), \Sigma_{\parsEnc}(\vecx^j_{Q^j}))||\mathcal{N}(\Z|\mu_{\parsEnc}(\vecx^j_{P^j}), \Sigma_{\parsEnc}(\vecx^j_{P^j})))
  $\;
  Compute joint ELBO \hfill%
    $\qquad\mathcal{L}_{\lambda,\parsDec,\parsEnc}= \,\mathcal{L}^{\elbo}_{\parsDec,\parsEnc}(\mathcal{B}_Q)-\lambda( r - \ell(\mathcal{B}_{\bar{P}}; \theta) 
 - \mathcal{L}^{\elbo}_{\parsDec,\parsEnc}(\mathcal{B}_P) + \mathcal{L}^{\elbo}_{\parsDec,\parsEnc}(\mathcal{B}_Q))$\;
  \tcc{Gradient descent step}
  Compute gradient  $g = \nabla_{\parsDec,\parsEnc} \mathcal{L}_{\lambda,\parsDec,\parsEnc}$ \;
  Use $g$ to update parameters $\parsDec$ and $\parsEnc$  \;
}
\end{algorithm2e}
}

\paragraph{Extension using normalizing flows.}
We can also use normalizing flows~\cite{rezende_flow_15} to enable more expressive approximate posterior distributions (see Appendix~\ref{sec:flow_background} for details on normalizing flows).
However, in this case, the KL divergences in Algorithm \ref{alg:cap} cannot be computed in closed form but have to be approximated, e.g., by sampling (see Appendix~\ref{section:flow_kl_approx}).

\section{Related Work}
\label{sec:related}

Because of space constraints, we only provide a limited treatment of related work here.
Further related work is discussed in Appendix~\ref{sec:extended-related}.

Generative models have a rich history of dealing with missing data 
\cite{jordan_em_imput1995,em_imputation_jordan1993,rubin2019}.
In recent years, also scalable deep generative models \cite{gan_2014_GF,vae13} have been considered for this setting, in particular GANs and VAEs.

\paragraph{GAN-based models.} GAN-based models are widely applied for recovering corrupted images due to their success in generating high-quality images \cite{dcgan,laplace_gan,face_completion_li,gan_face_allen,misgan}.
Furthermore, the GAIN model \cite{gain} was used for the imputation of missing data from the UCI repository  \cite{uci_data}. 
GAIN consists of a generator, which imputes data given partial data and a missingness mask, and a discriminator, which estimates a missingness mask from imputed data and hints of the mask.
A central challenge of applying GAN-based models for imputation is the difficulty of training them, i.e., solving a min-max optimization of nonlinear functions \cite{gan_tutorial17}. 

\paragraph{VAE-based models.}
In \cite{wu_pre_train_impute_vae}, corrupted images were imputed using a VAE model, which was pre-trained on fully observed training data.
Another approach, VAEs with zero imputation, was introduced in \cite{nazabal2020handling}, where the missing data is filled with zeros in the training and test stages and then fed to the VAE to obtain the imputation as the reconstruction of the decoder.
The imputation method MIWAE \cite{miwae} is based on importance-weighted autoencoders~\cite{iwae15} and also uses missing data imputed with zeros as an input to the inference network and utilizes a decoder for imputation.
Using zero-imputed data directly  as input to the encoder can lead to biased posterior estimates. Therefore, the VAE-PN/PNP models which use encoders based on the deep sets architecture~\cite{zaheer2017deep} as introduced in \cite{ma2019eddi} can be beneficial.
All the above approaches assume  MCAR or MAR data.
Recently, VAEs have also been considered for MNAR data \cite{vae_mnar_collier}.
Not-MIWAE \cite{ipsen2021notMIWAE} is another IWAE-based model  which considers the distribution of the missingness mechanism explicitly in the ELBO. 
Other approaches for dealing with the {MNAR} setting include  \cite{ident_mnar_iwae,deep_set_mnar,lim_iwae-mnar}.

\paragraph{Posterior consistency.}
A few papers have considered some form of posterior consistency for improving the performance of VAEs in different settings.
For instance, \cite{abstract_liu}  considered a notion of posterior consistency regarding augmentations of the input data.
In particular, the problem of inconsistency was addressed by regularizing the original ELBO objective with a weighted KL-divergence of latent variables encoded by real and decoded data.
In \cite{sinha2021consistency}, consistency of the posterior was required for the latent variables obtained from the original data and data under random transformation.
In \cite{zhu_nlp_reg_2022}, posterior consistency regularization for application in neural machine translation was enforced through the likelihood of the reconstruction $\hat{\vecx}$ using various data augmentation methods.
Although the aforementioned methods show promising results, they consider settings with fully observed data and the used approaches do not build on inherent properties that the posterior must satisfy like our work.

\section{Experimental Setup}
\label{sec:experimental-setup}

\subsection{Metrics}
To assess the advantages of our proposed regularization, we considered the following metrics in experiments: imputation quality and information curves.

\paragraph{Imputation quality.} 
To assess the quality of imputation we compute the root-mean-square error (RMSE) between te imputed and the ground-truth data, i.e., for a partially observed dataset $\calD=\{(\vecx^i_{Q^i})\}_{i=1}^n$ for which also the full observations are available we have
$
\textnormal{RMSE}(\calD) = ( \frac{1}{n} \sum_{i=1}^{n} \sum_{j \in \calV \setminus Q^i} (\hat{x}^i_j-\tilde{x}^i_j)^{2} )^{1/2},
$
where $\hat{\vecx}^i$ is the completed $i$th sample and $\tilde{\vecx}^i$ is the $i$th ground-truth sample.
Furthermore, the negative expected log-likelihood $
-\expectation_{\vecz \sim q_{\parsEnc}(\vecz| \vecx_Q)}[\log p_{\parsDec}(\tilde{\vecx}_{ \calV \setminus Q}| \vecz)]$, and ELBO values are considered to measure the quality of imputation. 

\paragraph{Information curve (IC).} 
Information curves allow us to investigate how useful the latent space of a VAE is for estimating the information gain of an unobserved variable.
The information curve regarding some target feature $x_{t}$ for sample $\vecx$ is computed as follows.
Starting from not observing any features at all, i.e., $O = \emptyset$, we iteratively select the next most informative feature $x_j$ regarding the target feature $x_{t}$ using an approximation of the information reward as in~\cite{ma2019eddi}:
\begin{align} \label{eq:reward_approx}
  j
  = \, & \underset{i \in U}{\arg \max }\, \mathbb{E}_{x_i \sim \hat{p}(X_i | \vecx_{O})} \kl\left[q(\Z | x_i, \vecx_{O}) \| q\left(\Z | \vecx_{O}\right)\right]- \\
 &\qquad\qquad \mathbb{E}_{x_t, x_i \sim \hat{p}\left(X_t, X_i | \vecx_{O}\right)} \kl \left[q(\Z | x_t, x_i,\vecx_{O}) \| q\left(\Z | x_t,\vecx_{O}\right)\right] \nonumber
\end{align}
where $U$ is the set of unobserved variables and $O$ is the set of observed variables.
The distribution $\hat{p}(X_i | \vecx_O) =\int_{\vecz} q_{\parsEnc}(\vecz | \x_O) p_{\parsDec}(X_i | \vecz) \textrm{d}\vecz$, where in practice the integration is approximated by samples. The distribution $\hat{p}(X_t, X_i|\vecx_O)$ is defined accordingly.
At each step, we measure the prediction quality regarding the target feature based on the available features, i.e., the squared error between $x_t$ and $\hat{x}_t$, where $\hat{x}_t$ is the prediction.
The errors over the iterations constitute the IC.

\subsection{Models, parameters, and model training}

\paragraph{Base models.} We consider the following baseline models:
\begin{enumerate*}[label=(\roman*), font=\itshape] 
  \item VAEs with zero-imputation for missing values (VAE-ZI),
  \item VAEs with zero-imputed missing values and an additional binary mask, indicating the available features, as input (Mask-VAE-ZI), 
  \item VAEs with the point-net-plus encoders as in \cite{ma2019eddi} (VAE-PNP), 
  \item MIWAEs~\cite{miwae}, and 
  \item Not-MIWAEs~\cite{ipsen2021notMIWAE}. Additionally, we also examine \item Flow-VAE, a variation of the partial VAE framework where the posterior distribution is approximated using normalizing flows.
\end{enumerate*}

\paragraph{Regularized models.}
For each baseline model we also consider its posterior-regularized version indicated by the prefix \textit{REG-}, e.g., REG-VAE-ZI is a posterior regularized variant of VAE-ZI.

\paragraph{Training with additional missingness.} 
For VAE-ZI, Mask-VAE-ZI, and VAE-PNP we also consider model training with artificial additional removal of observed features as proposed in \cite{ma2019eddi}.
The additional missingness is introduced at each iteration during the training process.
Features are randomly dropped with a rate of missingness sampled from $\mathcal{U}(0, 0.7)$.
Models trained with such additional missingness are indicated by the postfix \textit{-AM}.

\paragraph{Model architectures.}
The architectures of the models and their parameters were taken from the original papers, except for MIWAE and Not-MIWAE, where different scaling of data and activation functions were used. The reason for this choice was the inability to reproduce the same results as in the original setting; therefore, a couple of changes were made to boost the performance of the models. 
For flow-based partial VAEs, we used the piecewise-linear coupling transform~\cite{muller_neural_import} to increase the expressiveness of the approximate posterior distribution. 

\paragraph{Model training.}
All models were trained for 3000 epochs using ADAM~\cite{adam} with a learning rate of $0.001$ and a batch size of $64$ samples.
The data was scaled to a range from 0 to 1.
The parameters for the regularized VAEs, $\mathcal{P}$ and $\lambda$, were tuned by the imputation quality performance of the model on the training data.
For a detailed description of the parameters used for model training see Appendix~\ref{sec:detailed_parameters}.

\subsection{Data and missing values}
\label{sec:data}

In our experiments, we considered data from the UCI repository~\cite{uci_data}.
In particular, in line with previous work, we considered the following datasets: \emph{Boston housing}, \emph{Wine}, \emph{enb}, \emph{Breast cancer}, \emph{Yeast}, and \emph{Concrete}.

For the MCAR setting, the mask indicating missing values (missingness mask) was randomly sampled at the beginning of each run with 30, 50, and 70 \% of missingness.
For the MNAR setting, we considered self-censoring in which the missingness was created based on the mean value of a feature: a value is missing if it is higher than its mean.
Consequently, the mask has a fixed missing rate.

\section{Experiments}
\label{sec:experiments}

In this section, we empirically demonstrate the advantages of our proposed regularization for a variety of models.
Because of space constraints, we only highlight a selection of empirical results. See appendices~\ref{section:remaining_partial_experiments},~\ref{sec:flow_with_mnar}, and~\ref{sec:experiments-full-data} for additional results.

\subsection{Imputation Results}
We first investigate the effect of our regularization on the imputation quality.
We compute the RMSEs (Tables~\ref{table:general_comparison_impute_qual},~\ref{table:flow_comparison},~\ref{table:not_miwae_comparison} and Appendix~\ref{section:additional_rmse}), ELBOs (Appendix~\ref{sec:exp-additional}) and negative log-likelihoods (Appendix~\ref{section:additional_nllh}) on the test set, which is selected randomly for each run and contains 10 $\%$ of the data. 
Each experiment was repeated $10$ times to compute the statistics over these runs except for the MIWAE and Not-MIWAE models for which experiments were repeated $5$ times. %

Our first set of results for comparing different types of models with respect to RMSE is presented in Table \ref{table:general_comparison_impute_qual}.
We observe that the models with consistency regularization significantly outperform those with and without additional missingness (-AM), demonstrating the advantage of additionally considering the relationship between posteriors for $\vecx_Q$ and $\vecx_P$. For results with higher rates of missingness see Appendix~\ref{section:remaining_partial_experiments}---the results are qualitatively similar.
It should be noted that AM \cite{ma2019eddi} can improve the performance of the VAE-ZI and Mask-VAE-ZI models for some datasets in terms of imputation quality.
However, using AM alone is not sufficient because it can be disadvantageous for some partial VAE models, such as VAE-PNP in our experiments.
In contrast to vanilla AM, our regularization method is advantageous for all classes of considered VAE models.

In Table~\ref{table:flow_comparison}, we present results for applying the proposed regularization on VAEs with flows.
We observe that consistency regularization is advantageous for partial VAEs with normalizing flows in terms of imputation quality as well (although improvements are not that large), thereby demonstrating the flexibility of the proposed method. Additionally, it should be noted that partial VAEs with normalizing flows can deliver satisfactory results compared to other models for imputation in the MNAR setting.
Experimental results for Flow-VAEs in the MNAR setting are presented in Appendix~\ref{sec:flow_with_mnar}.

Furthermore, we evaluated the quality of imputation in a simple MNAR setting using the Not-MIWAE model. Following the pipeline of the authors, the model was trained and evaluated on the entire dataset, without splitting. Our results are presented in Table \ref{table:not_miwae_comparison} and we can observe that our regularization improves the imputation quality in the MNAR setting, even though the artificial missingness added for generating $\vecx_P$ is MCAR.

Additionally, different missing mechanisms for $\vecx_P$ were examined, cf. Appendix~\ref{sec:experiment_p_mechanisms}. We can also observe improvements in terms of imputation quality for some models for other types of missingness mechanisms. %
Furthermore, we considered the dependence of imputation quality on the probability of removing features $\mathcal{P}$, cf.\ Algorithm~\ref{alg:cap}, in Appendix \ref{sec:experiment_rmse_over_p_missing}. The experiment revealed that the optimal value of $\mathcal{P}$ closely aligns with the actual missing rate. %

Moreover, we tested our regularization on VAEs trained on fully observed data.
As shown in Appendix \ref{sec:experiments-full-data}, regularization provides a slight improvement in terms of the reconstruction quality, which demonstrates the advantage of this method even for the fully observed case.

\begin{table}[!tbp]
\caption{Imputation quality (RMSE) for 30~\% missingness. Results computed on the test data. Smaller is better.}
\label{table:general_comparison_impute_qual}
\centering
\begin{tabular}{cccc}
\toprule
 & \multicolumn{3}{c}{Dataset} \\\cmidrule{2-4}
 Model & Housing & Wine & enb \\\midrule
 VAE-ZI-AM & $0.1967 \pm 0.0050$ & $0.1289 \pm 0.0019$ & $0.2754 \pm 0.0072$ \\
 VAE-ZI    & $0.1976 \pm 0.0066$ & $0.1265 \pm 0.0022$ & $0.2846 \pm 0.0081$ \\
 REG-VAE-ZI & $\mathbf{0.1874\pm 0.0048}$ & $\mathbf{0.1238 \pm 0.0020 }$ & $\mathbf{0.2596 \pm 0.0057 }$ \\[1mm]
 Mask-VAE-ZI-AM & $0.1863 \pm 0.0047 $ & $0.1278 \pm 0.0019$ & $0.2611 \pm  0.0048$ \\
 Mask-VAE-ZI & $0.1892 \pm 0.0036$ & $0.1272 \pm 0.0021$ & $0.2560 \pm  0.0045$ \\
 REG-Mask-VAE-ZI & $\mathbf{0.1758\pm 0.0060}$ & $\mathbf{0.1235 \pm 0.0017 }$ & $\mathbf{0.2471 \pm 0.0042 }$  \\[1mm]
 VAE-PNP-AM & $0.1861 \pm 0.0055$  & $0.1300 \pm 0.0023$ & $0.2698 \pm 0.0074 $ \\
 VAE-PNP & $0.1837 \pm 0.0055$ & $0.1272 \pm 0.0020$ & $0.2592 \pm 0.0055$ \\
 REG-VAE-PNP & $\mathbf{0.1739 \pm 0.0044}$ & $\mathbf{0.1245 \pm 0.0019 }$ & $\mathbf{0.2435 \pm 0.0049 }$ \\
\bottomrule
\end{tabular}
\end{table}

\begin{table*}[tbp]
\caption{Imputation quality (RMSE) of VAE-flow, REG-VAE-flow, MIWAE, and REG-MIWAE for 30~\% missingness on test data. Smaller is better.}
\label{table:flow_comparison}
\centering
\scriptsize
\begin{tabular}[t]{cccc}
\toprule
Dataset & VAE-flow & REG-VAE-flow \\\cmidrule{1-3}
Housing & $0.1697 \pm 0.0064$ & $\mathbf{0.1627 \pm 0.0058}$  \\
Wine & $0.1085 \pm 0.0022$ & $\mathbf{0.1067 \pm 0.0021}$ \\
enb  & $ 0.2069 \pm 0.0034 $ & $\mathbf{ 0.2031 \pm 0.0034}$  \\
\bottomrule
\end{tabular}%
\hspace{3mm}
\begin{tabular}[t]{cccc}
\toprule
Dataset &  MIWAE & REG-MIWAE \\\cmidrule{1-3}
Red Wine & $  0.1452 \pm 0.0028 $ & $\mathbf{0.1205 \pm  0.0015 }$ \\
concrete & $ 0.2755 \pm  0.0146$ & $\mathbf{0.2046 \pm  0.0046 }$ \\
White Wine & $ 0.1208 \pm 0.0041 $ & $\mathbf{0.0996 \pm  0.0014 }$ \\
banknote & $ 0.3312 \pm 0.0332 $ & $\mathbf{0.2715 \pm  0.0087 }$ \\
breast & $  0.1545 \pm 0.0069 $ & $\mathbf{0.0906 \pm  0.0037 }$ \\
yeast & $  \mathbf{0.1098 \pm 0.0050} $ & $0.1197 \pm  0.0036 $ \\
\bottomrule
\end{tabular}
\end{table*}

\begin{table}[tbp]
\sisetup{detect-weight=true,detect-inline-weight=math,separate-uncertainty=true}

\caption{Imputation quality (RMSE) of Not-MIWAE and REG-Not-MIWAE.}
\label{table:not_miwae_comparison}
\centering
\begin{tabular}{cS[table-format=1.4(4)]S[table-format=1.4(4)]}
\toprule
Dataset 		&  \textnormal{Not-MIWAE} & \textnormal{REG-Not-MIWAE} \\\cmidrule{1-3}
Red Wine 		& 0.1594(0.0225) & \bfseries 0.1269(0.0256) \\
concrete 		& 0.2887(0.0387) & \bfseries 0.2558(0.0419) \\
White Wine 		& 0.0891(0.0100) & \bfseries 0.0842(0.0134) \\
banknote 		& 0.2459(0.0269) & \bfseries 0.1682(0.0301) \\
breast 			& 0.1000(0.0012) & \bfseries 0.0632(0.0017) \\
yeast 			& 0.1363(0.0010)  & \bfseries 0.1351(0.0008) \\
\bottomrule
\end{tabular}
\end{table}

\begin{figure*}[!tb]
  \centering
  \setlength{\belowcaptionskip}{-0.2\baselineskip}
  \begin{subfigure}[c]{0.32\textwidth}
    \includegraphics[width=\textwidth]{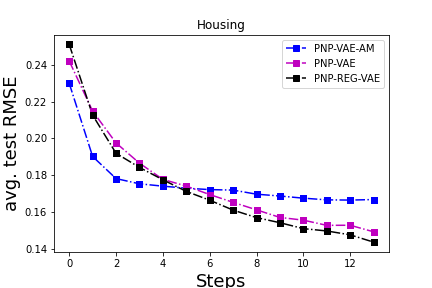}
    \subcaption{Boston housing}
  \end{subfigure}
  \begin{subfigure}[c]{0.32\textwidth}
    \includegraphics[width=\textwidth]{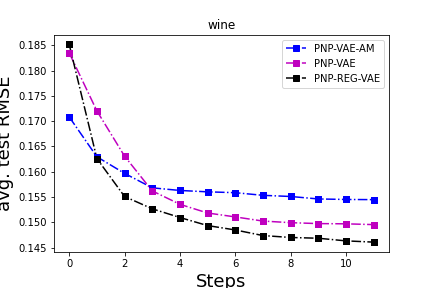}
    \subcaption{wine}
  \end{subfigure}
  \begin{subfigure}[c]{0.32\textwidth}
    \includegraphics[width=\textwidth]{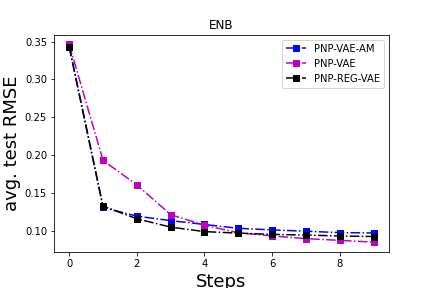}
    \subcaption{enb}
  \end{subfigure}
  \caption{Information curves on test data with 30~\% missingness for partial VAEs. %
  }
  \label{fig:pnp_comparison_main}
\end{figure*}

\subsection{Active Feature Acquisition Results}
In this experiment, we aimed to evaluate the effect of regularization on the efficiency of information acquisition. All models were initially trained on partial training datasets.
We ran the IC experiment ten times on the test set.

Our results are presented in Figure~\ref{fig:pnp_comparison_main}.
We can observe that partial VAEs with AM achieve better test RMSE scores than other models in the early steps for some datasets, indicating that a small number of features acquired is utilized more effectively. However, as soon as a sufficient number of features are available, AM generally worsens the performance of the model, and the base partial VAE delivers better results.
Partial VAEs with consistency regularization outperform all partial VAEs with AM in terms of the test RMSE after the first couple of steps, whereby in contrast with AM almost all base partial VAEs are outperformed during the entire procedure. The results demonstrate the importance of consistency regularization for efficient information acquisition.

In addition, our regularization can increase the efficiency of information acquisition for VAE models equipped with normalizing flows, although improvements are typically much smaller, cf.\ Appendix~\ref{sec:additional_ICs} for details.

\section{Conclusions}
\label{sec:conclusions}
We have considered the challenge of learning VAEs from incomplete data.
In particular, we focused on improving the amortized approximate posterior distributions regarding the missingness in the data.
To this end, we formalized a notion of posterior consistency with respect to the missingness and proposed a regularizer that improves a VAE's encoder posterior consistency when used during training.
We showed that using our proposed regularization improves the imputation quality for different classes of VAEs and different types of missingness.
Finally, we showed that our regularized VAEs often outperform VAEs without regularization on downstream tasks leveraging the latent space of the VAE for making decisions.
We believe that our paper can improve the usage of VAEs in many practical settings on partial data.
In future work, we will analyze the regularization of partial VAEs for MNAR settings in more detail.

\bibliographystyle{splncs04}
\bibliography{refs}

\clearpage
\appendix

\onecolumn

\section{Posterior consistency: Proof of Observation~\ref{obs:posterior}}
\label{sec:posterior-consistency-observation}

Let $\vecx$ be a sample with missing features and let $Q$ denote a subset of the features of $\vecx$ that is not missing.

\paragraph{MCAR data.} 
For MCAR we have
\begin{align}
  p(\vecz| \vecx_Q) &= p(\vecz| \vecx_Q, \vecx_{\calV - Q} = \perp) \nonumber\\
     &\myeq{(a)} \frac{p(\vecx_Q, \vecx_{\calV - Q} = \perp | \vecz) p(\vecz)}{p(\vecx_Q, \vecx_{\calV - Q} = \perp)} \nonumber\\
    &= \frac{p(\vecx_P, \vecx_{\bar{P}}, \vecx_{\calV - Q} = \perp| \vecz) p(\vecz)}{p(\vecx_Q, \vecx_{\calV - Q} = \perp)} \notag\\
    &= \frac{p(\vecx_P | \vecx_{\bar{P}}, \vecx_{\calV - Q} = \perp, \vecz) p(\vecx_{\bar{P}}, \vecx_{\calV - Q} = \perp | \vecz) p(\vecz)}{p(\vecx_Q, \vecx_{\calV - Q} = \perp)}\notag \\
    &\myeq{(b)} \frac{p(\vecx_P | \vecz) p(\vecx_{\bar{P}}, \vecx_{\calV - Q} = \perp | \vecz) p(\vecz)}{p(\vecx_Q, \vecx_{\calV - Q} = \perp)}\notag  \\
    &\myeq{(c)} \frac{p(\vecx_P | \vecz) p(\vecx_{\bar{P}} | \vecz) p(\vecx_{\calV - Q} = \perp) p(\vecz)}{p(\vecx_Q) p(\vecx_{\calV - Q} = \perp)} \notag  \\
    &= \frac{p(\vecx_{\bar{P}} | \vecz) p(\vecz | \vecx_{P}) p(\vecx_{P}) p(\vecz)}{p(\vecx_Q) p(\vecz)} \notag\\
    &= \Big[ \frac{p(\vecx_{P})}{p(\vecx_Q)} p(\vecx_{\bar{P}} | \vecz) \Big] p(\vecz| \vecx_P) , \nonumber
\end{align}
where \textit{(a)} is because in MCAR the missingness is independent of the values in $\tilde{\vecx}$, \textit{(b)} is also because of the MCAR assumption and the assumption that the dimension of $\tilde{\vecx}$ are generated independently given $\vecz$, and \textit{(b)} is because of the MCAR assumption.

\paragraph{MNAR data.} 
For MNAR data we have
\label{sec:mnar}
\begin{align*}
    p(\vecz| \vecx_Q, \vecx_{\calV - Q} = \perp) &= \frac{ p(\vecz,\vecx_Q, \vecx_{\calV - Q} = \perp) }{p(\vecx_Q, \vecx_{\calV - Q} = \perp)} \\
    &= \frac{ p(\vecz,\vecx_P, \vecx_{\bar{P}}, \vecx_{\calV - Q} = \perp) }{p(\vecx_Q, \vecx_{\calV - Q} = \perp)}\\
    &= p\left(\vecz \mid \vecx_P, \vecx_{\calV-Q}=\perp \right) \cdot p\left(\vecx_{\bar{P}}\mid \vecz, \vecx_P, \vecx_{\calV-Q}=\perp \right) \cdot \\
    &\qquad \frac{p\left(\vecx_P, \vecx_{\calV-Q}=\perp \right)}{p\left(\vecx_Q,  \vecx_{\calV-Q}=\perp\right)}
\end{align*}

\section{Posterior regularization}
\label{sec:reg_derivation}
Including the posterior consistency requirement in the VAE framework is equivalent to solving the following optimization problem
\begin{align} \label{eq:optim_pr}
    \max_{\theta, \phi}& \quad \mathcal{L}^{\elbo}(\vecx_Q^i) \\
    \textnormal{ s.t.} & \quad \kl \Big(q_\phi(\Z \mid \vecx_Q^i) \| \frac{p_{\mathcal{\theta}}(\vecx_P^i)}{p_{\mathcal{\theta}}(\vecx_Q^i)} \cdot p_\theta(\vecx_{\bar{P}}^i \mid \Z) q_\phi(\Z \mid \vecx_P^i)\Big)= 0 \nonumber \\
    &\quad \forall P \subseteq Q \subseteq \calV, \nonumber
\end{align}
where data point $i$ is considered.
Please that the above optimization problem might not be feasible.

This optimization problem can be rewritten in terms of the Lagrangian function for a fixed subset $P_j \subseteq Q$: 
\begin{align} \label{eq:lag_func}
    \mathcal{L}_{\lambda, \theta, \phi} &= \mathcal{L}_{\theta, \phi}^{\elbo}(\X_Q^i) -\lambda_j  \kl \Big(q_\phi(\Z | \X_Q^i) \| \frac{p_{\mathcal{\theta}}(\X_{P_j}^i)}{p_{\mathcal{\theta}}(\X_Q^i)} \cdot p_\theta(\X_{\bar{P}_j}^i | \Z) q_\phi(\Z \mid \X_{P_j}^i)\Big) \nonumber \\
    &=  \mathcal{L}_{\theta, \phi}^{\elbo}(\X_Q^i) - \lambda_j \Big[\kl(q_\phi(\Z | \X_Q^i) \| q_\phi(\Z | \X_{P_j}^i)) - \nonumber\\
        &\qquad\qquad \mathbb{E}_{q_{\phi}(\Z | \X_Q^i)} \log p_\theta(\X_{\bar{P}_j}^i | \Z)  - \mathbb{E}_{q_{\phi}(\Z | \X_Q^i)} \log \frac{p_{\mathcal{\theta}}(\X_{P_j}^i)}{p_{\mathcal{\theta}}(\X_Q^i)} \Big] \nonumber \\
    &= \mathcal{L}_{\theta, \phi}^{\elbo}(\X_Q^i) - \lambda_j \Big[\kl(q_\phi(\Z | \X_Q^i) \| q_\phi(\Z | \X_{P_j}^i)) - \nonumber\\
        &\qquad\qquad \mathbb{E}_{q_{\phi}(\Z | \X_Q^i)} \log p_\theta(\X_{\bar{P}_j}^i | \Z) - \log \frac{p_{\mathcal{\theta}}(\X_{P_j}^i)}{p_{\mathcal{\theta}}(\X_Q^i)}\Big] \nonumber \\
    &\approx \mathcal{L}_{\theta, \phi}^{\elbo}(\X_Q^i) - \lambda_j \Big[\kl(q_\phi(\Z | \X_Q^i) \| q_\phi(\Z | \X_{P_j}^i)) - \nonumber\\
        &\qquad\qquad \mathbb{E}_{q_{\phi}(\Z | \X_Q^i)} \log p_\theta(\X_{\bar{P}_j}^i | \Z)  - (  \mathcal{L}_{\theta, \phi}^{\elbo}(\X_P^j)- \mathcal{L}_{\theta, \phi}^{\elbo}(\X_Q^i))\Big], \nonumber
\end{align}
where $\log \tfrac{p_{\mathcal{\theta}}(\X_{P_j}^i)}{p_{\mathcal{\theta}}(\X_Q^i)}$ is approximated through the difference of ELBO terms. 
The quality of this approximation depends on the class of VAE model used.
Models with more expressive posteriors, i.e., flows, typically have smaller inference gap \cite{cremer2018inference}, which leads to a better approximation. \\
The gradient of the Lagrangian function has then to be computed w.r.t.\ $\theta$, $\phi$ and $\lambda$, i.e.
\begin{equation} \label{eq:lag_func}
    \nabla_{\theta, \phi, \lambda}\mathcal{L}_{\lambda, \theta, \phi} =
    \left(\nabla_{\theta}\mathcal{L}_{\lambda, \theta, \phi}, \nabla_{\phi}\mathcal{L}_{\lambda, \theta, \phi}, \nabla_{\lambda}\mathcal{L}_{\lambda, \theta, \phi} \right) \\
\end{equation}
and solved for
\begin{equation} \label{eq:lag_func}
    \nabla_{\theta, \phi, \lambda}\mathcal{L}_{\lambda, \theta, \phi} =
    0.
\end{equation}

Following the above considerations, to (approximately) enforce consistency in the VAE's approximate posterior, we regularize the ELBO for each fixed subset $P_j \subseteq Q$ as follows
\begin{align} \label{eq:elboext}
 \mathcal{L}_{\lambda, \theta, \phi} &= \mathcal{L}_{\theta, \phi}^{\elbo}(\X_Q^i) - \lambda_j \Big[ \kl(q_\phi(\Z | \X_Q^i) \| q_\phi(\Z | \X_{P_j}^i)) - \\
    &\qquad\qquad \mathbb{E}_{q_{\phi}(\Z | \X_Q^i)} \log p_\theta(\X_{\bar{P}_j}^i | \Z) - (  \mathcal{L}_{\theta, \phi}^{\elbo}(\X_P^j)- \mathcal{L}_{\theta, \phi}^{\elbo}(\X_Q^i)) \Big]. \nonumber
\end{align}

\section{Background on normalizing flows}
\label{sec:flow_background}
In many cases, using a normal distribution with diagonal covariance as the variational family is too restrictive, i.e., it cannot represent the true posterior. 
This is in particular the case when working with missing data: even if the posterior distribution for complete data points can be accurately represented by a standard normal distribution this might not be the case if data is only partially observed.

Normalizing flows~\cite{rezende_flow_15} enable more expressive posterior distributions.
The key idea of normalizing flows is to transform a sample $\vecz' \in \mathbb{R}^k$ from a (simple) base distribution $\tilde{p}(\Z')$ through a series of invertible transformations $f_1, \ldots, f_l$, where each $f_i\colon \mathbb{R}^k \rightarrow \mathbb{R}^k$, such that $\vecz = f(\vecz') = f_l(f_{l-1}(\cdots(f_1(\vecz'))))$.
Then,
\begin{align}
    \log p(\vecz) = \log \tilde{p}(f^{-1}(\vecz)) + \sum_{i=1}^l \log \left| \det \frac{\partial f_i^{-1}(\vecz)}{\partial \vecz} \right|,
\end{align}
where $\det$ denotes the determinant and $f_i^{-1}$ is the inverse of $f_i$.
For suitable choices of the functions $f_i$ these determinants can be computed efficiently, enabling the computation of the probability of the transformed sample.
The expressiveness of normalizing flows depends on the functions $f_i$ (e.g., see~\cite{kong2020expressive}), and several functions have been proposed therefore (e.g., see~\cite{kobyzev2020normalizing} for an overview).

\section{Approximation of KL-divergence term}
\label{section:flow_kl_approx}
The KL-divergence terms from Equation~\ref{eq:reward_approx} for VAE-Flow were approximated using log-likelihood ratio
 \begin{align*} %
 \kl(q_\parsEnc(\Z | \vecx_Q^i) \| q_\parsEnc(\Z | \vecx_{P^i}^i)) &= 
  \mathbb{E}_{\vecz \sim q_\parsEnc(\vecx^i_{Q_i})} \log \frac{q_\parsEnc(\Z | \vecx^i_{Q^i})}{q_\parsEnc(\Z | \vecx^i_{P^i})} 
   \approx \sum_{m=1}^M \log \frac{q_\parsEnc(\vecz_m | \vecx^i_{Q^i})}{q_\parsEnc(\vecz_m | \vecx^i_{P^i})},
 \end{align*}
where $M$ is the number of samples used in the approximation and $\vecz_m \sim q_\parsEnc(\vecx^i_{Q_i})$.

\section{Extended Related Work}
\label{sec:extended-related}

Generative models have a rich history of applications in the domain of missing data 
\cite{jordan_em_imput1995,em_imputation_jordan1993,rubin2019}. 
In recent years, also highly scalable deep generative models \cite{gan_2014_GF,vae13} have been introduced.
Deep generative models for data imputation can be broadely divided into two categories, \emph{VAE-based} and \emph{GAN-based} models.

\paragraph{GAN-based models.}
GAN-based models are commonly applied for recovering corrupted images because of their success in generating high-quality images \cite{dcgan,laplace_gan}.
Early approaches, \emph{Generative Face Completion} \cite{face_completion_li} and \emph{Generative Adversarial Denoising Autoencoder for Face Completion} \cite{gan_face_allen}, dealt with the completion of corrupted face images from the CelebA dataset \cite{celeba}.
Both approaches are represented as a mixture of Autoencoder and GAN architectures.
In order to be able to impute corrupted images, these models have to be trained on the fully observed training dataset, which is not applicable in the real world.
The \emph{GAIN} model \cite{gain} was applied for the imputation of UCI datasets  \cite{uci_data} but can also be used for images \cite{misgan}. 
\emph{GAIN} consists of a generator, which imputes data given partial data and a missing mask, and a discriminator, which estimates a missing mask from imputed data and hints of the mask. GAIN outperformed many state-of-the-art methods for imputation on UCI datasets such as: \emph{MICE} \cite{mice}, \emph{MissForest} \cite{missforest} and \emph{Matrix completion} \cite{JMLR:v11:mazumder10a}.
Another method for imputation of images, \emph{MisGAN} \cite{misgan}, is represented through a more complex architecture, where the data and mask matrix have their own generator and discriminator. For the imputation of the data, an additional generator and discriminator are introduced during training.
The results of image imputation with a high percentage of missingness are impressive and surpass all baseline methods, such as GAIN and Matrix completion.
A big issue of many GAN-based models for imputation is the complexity of training such models, where training requires (approximately) solving a min-max optimization of nonlinear functions and the requirement of large training datasets

\paragraph{VAE-based models.}
In \cite{wu_pre_train_impute_vae}, corrupted images were imputed using a VAE model, which was pre-trained on fully observed training data--a setting that is of limited relevance for many real-world problems.
Another approach, \emph{VAE with Zero Imputation
} (VAE-ZI), was introduced in \cite{nazabal2020handling}, where the missing data is filled with zeros in the training and test stages and then fed to the VAE to obtain imputation as a reconstruction of the decoder.
The imputation method \emph{MIWAE} \cite{miwae}, which is based on \emph{IWAE} \cite{iwae15}, also uses missing data imputed with zeros as an input to the inference network and utilizes a decoder for imputation.
The proposed method is expected to be better for imputation because of its lower bound \cite{iwae15}, which becomes tighter as the number of importance weights increases.
The model can be very complex, depending on the dataset, as a consequence of the architecture. 
Using zero-imputed data directly as input to the encoder can lead to biased posterior estimation.
Therefore, the \emph{VAE-PN/PNP} approach was introduced in \cite{ma2019eddi} which approaches this problem using a permutation-invariant input of the observed data to the inference network.
All the above approaches assume MCAR or MAR patterns of missingness. 
Recently, VAE-based models were proposed for handling MNAR missingness.
In \cite{vae_mnar_collier} the lower bound of the observed data is maximized given a mask matrix to impute data with the MNAR missingness.
The \emph{IWAE}-based model from \cite{ipsen2021notMIWAE}, \emph{Not-MIWAE}, considers the distribution of the missing mechanism explicitly in ELBO given reconstruction from the decoder.
Other methods for dealing with the {MNAR} setting include  \cite{ident_mnar_iwae,deep_set_mnar,lim_iwae-mnar}.
Nevertheless, none of the above proposed models consider the consistency of the posterior distribution, which is crucial for learning the correct model of the data.
We propose a consistency regularization method, which is suitable for all mentioned VAE-based models, showing experimentally its advantage on a selected set of models.
A similar method for dealing with partial data was introduced in \cite{carvalho_arbitrary}, in which the KL-divergence of posteriors encoded by full and partial data was used during the training of partial VAE.
However, initially, the VAE model must be trained on fully observed data to use its inference network during the training of the partial VAE. %
Using the same idea, our method allows us to train partial {VAE} and compute the KL-divergence of two posteriors simultaneously.
The consistency of the posterior has also been the subject of investigations orthogonal to missingness, e.g., in \cite{abstract_liu} consistency was defined as the ability of the inference network to match the true posterior of the generative model.
To address the problem of inconsistency, regularizing the original ELBO objective with a weighted KL-divergence of latent variables encoded by real and reconstructed data was proposed by \cite{abstract_liu}.
In \cite{sinha2021consistency}, to enforce the consistency of the posterior, the ELBO was regularized with the KL-divergence of latent variables encoded by original data and data under random transformation.
In \cite{zhu_nlp_reg_2022}, posterior consistency regularization was used for application in neural machine translation. One of the many differences with the other mentioned methods is the use of multiple data augmentation methods, where one of the augmentation methods coincides with our artificial missingness for $\vecx_P $, created uniformly at random. 
Although the aforementioned methods show impressive results, they only consider settings with fully observed data.
In this study, we aimed to address the consistency issue by using partial data.

\section{Additional Experimental Details}
\label{sec:detailed_parameters}
For VAE-ZI, Mask-VAE-ZI, and VAE-PNP, experimental settings, architectures, and parameters were taken from \cite{ma2019eddi}, where in the case of MIWAE and Not-MIWAE, a few things were different from those in the original papers \cite{miwae,ipsen2021notMIWAE}, scaling of the data was changed to $[0,1]$ range, and the activation function for the Decoder was changed to a sigmoid function. The mapping $\pi_{\gamma, j}(\vecx)$ for the Not-MIWAE model was computed using self-masking.\\
The Flow-VAE model has the following architecture and parameters for all UCI datasets: the Encoder is a $2D$-$100$-$100$-$100$-$K$ neural network with an ELU activation function that maps the concatenation of the data sample and missing mask to the latent space, and the Decoder is a $K$-$100$-$100$-$100$-$100$-$D$ neural network with an ELU activation function that reconstructs data from the latent space, where $D$ denotes a data dimension and $K$ represents a latent dimension (10 for this model). Number of bins used for Piecewise-Linear Coupling Transform is 10. Experimental settings for Flow-VAE are taken as for models from \cite{ma2019eddi}.\\
For the MCAR setting, the missing mask was uniformly created at the beginning of each run at random with 30, 50 and 70 percents of missingness for the VAE-ZI, Mask-VAE-ZI, VAE-PNP, and Flow-VAE models. For the MIWAE model, a missing mask  with a 50 percent of missingness was created for each run.

The optimal consistency regularization parameters were selected based on their imputation quality performance on the training data. Tables \ref{table:lambda_optimal_reg_vae_zi}, \ref{table:p_optimal_reg_vae_zi}, \ref{table:lambda_optimal_reg_mask_vae_zi}, \ref{table:p_optimal_reg_mask_vae_zi}, \ref{table:lambda_optimal_reg_vae_pnp}, and \ref{table:p_optimal_reg_vae_pnp}, list the optimal parameters for REG-VAE-ZI, REG-Mask-VAE-ZI, and REG-PNP-VAE on data with 30, 50, and 70 $\%$ of missingness. Meanwhile, Tables \ref{table:params_optimal_reg_miwae} and \ref{table:params_optimal_reg_not_miwae} present the optimal parameters for REG-MIWAE and REG-Not-MIWAE. The optimal parameters for REG-Flow-VAE on data with 30 $\%$  of missingness are listed in Table \ref{table:params_optimal_reg_vae_flow}. The optimal $\lambda$ parameters were selected from the range $[0.01, 1.5]$ and $\mathcal{P}$ parameters were selected from the range $[0.01, 0.8]$.

\begin{table*}[!tbp]
\caption{Optimal $\lambda$ parameters for REG-VAE-ZI.}
\label{table:lambda_optimal_reg_vae_zi}
\centering
\begin{tabular}{ccccc}
\toprule
 & \multicolumn{3}{c}{REG-VAE-ZI} \\\cmidrule{2-4}
Dataset & $30\%$ missingness & $50\%$ missingness & $70\%$ missingness \\\cmidrule{1-4}
Housing  & $1.0$ &  $1.5$ & $1.5$  \\
Wine &  $1.5$  & $1.5$ & $1.5$ \\
enb & $1.5$  &  $1.5$ & $0.5$  \\
\bottomrule
\end{tabular}
\end{table*}

\begin{table*}[!tbp]
\caption{Optimal $\mathcal{P}$ parameters for REG-VAE-ZI.}
\label{table:p_optimal_reg_vae_zi}
\centering
\begin{tabular}{ccccc}
\toprule
 & \multicolumn{3}{c}{REG-VAE-ZI} \\\cmidrule{2-4}
Dataset & $30\%$ missingness & $50\%$ missingness & $70\%$ missingness \\\cmidrule{1-4}
Housing  & $0.3$ &  $0.5$ & $0.6$  \\
Wine &  $0.4$ &  $0.6$ & $0.6$ \\
enb & $0.3$ &  $0.3$ & $0.3$  \\
\bottomrule
\end{tabular}
\end{table*}

\begin{table*}[!tbp]
\caption{Optimal $\lambda$ parameters for REG-Mask-VAE-ZI.}
\label{table:lambda_optimal_reg_mask_vae_zi}
\centering
\begin{tabular}{ccccc}
\toprule
 & \multicolumn{3}{c}{REG-Mask-VAE-ZI} \\\cmidrule{2-4}
Dataset & $30\%$ missingness & $50\%$ missingness & $70\%$ missingness \\\cmidrule{1-4}
Housing  & $1.0$ &  $0.5$ & $1.5$  \\
Wine &  $1.5$  & $1.5$ & $1.5$ \\
enb & $1.5$  &  $1.5$ & $1.5$  \\
\bottomrule
\end{tabular}
\end{table*}

\begin{table*}[!tbp]
\caption{Optimal $\mathcal{P}$ parameters for REG-Mask-VAE-ZI.}
\label{table:p_optimal_reg_mask_vae_zi}
\centering
\begin{tabular}{ccccc}
\toprule
 & \multicolumn{3}{c}{REG-Mask-VAE-ZI} \\\cmidrule{2-4}
Dataset & $30\%$ missingness & $50\%$ missingness & $70\%$ missingness \\\cmidrule{1-4}
Housing  & $0.3$ &  $0.4$ & $0.5$  \\
Wine &  $0.3$ &  $0.6$ & $0.7$ \\
enb & $0.4$ &  $0.4$ & $0.4$  \\
\bottomrule
\end{tabular}
\end{table*}

\begin{table*}[!tbp]
\caption{Optimal $\lambda$ parameters for REG-VAE-PNP.}
\label{table:lambda_optimal_reg_vae_pnp}
\centering
\begin{tabular}{ccccc}
\toprule
 & \multicolumn{3}{c}{REG-VAE-PNP} \\\cmidrule{2-4}
Dataset & $30\%$ missingness & $50\%$ missingness & $70\%$ missingness \\\cmidrule{1-4}
Housing  & $0.5$ &  $1.0$ & $1.5$  \\
Wine &  $1.0$  & $1.5$ & $1.5$ \\
enb & $1.5$  &  $1.5$ & $1.5$  \\
\bottomrule
\end{tabular}
\end{table*}

\begin{table*}[!tbp]
\caption{Optimal $\mathcal{P}$ parameters for REG-VAE-PNP.}
\label{table:p_optimal_reg_vae_pnp}
\centering
\begin{tabular}{ccccc}
\toprule
 & \multicolumn{3}{c}{REG-VAE-PNP} \\\cmidrule{2-4}
Dataset & $30\%$ missingness & $50\%$ missingness & $70\%$ missingness \\\cmidrule{1-4}
Housing  & $0.1$ &  $0.5$ & $0.7$  \\
Wine &  $0.4$ &  $0.4$ & $0.7$ \\
enb & $0.5$ &  $0.6$ & $0.5$  \\
\bottomrule
\end{tabular}
\end{table*}

\begin{table*}[!tbp]
\caption{Optimal parameters for REG-Flow-VAE on different datasets with $30\%$ missingness.}
\label{table:params_optimal_reg_vae_flow}
\centering
\begin{tabular}{ccc}
\toprule
 & \multicolumn{2}{c}{REG-Flow-VAE} \\\cmidrule{2-3}
Dataset & $\lambda$ & $\mathcal{P}$ \\\cmidrule{1-3}
Housing  & $0.5$ &  $0.6$ \\
Wine &  $0.5$ &  $0.3$ \\
enb & $0.1$ &  $0.1$ \\
\bottomrule
\end{tabular}
\end{table*}

\begin{table*}[!tbp]
\caption{Optimal parameters for REG-MIWAE on different datasets with $50\%$ missingness.}
\label{table:params_optimal_reg_miwae}
\centering
\begin{tabular}{ccc}
\toprule
 & \multicolumn{2}{c}{REG-MIWAE} \\\cmidrule{2-3}
Dataset & $\lambda$ & $\mathcal{P}$ \\\cmidrule{1-3}
White Wine  & $1.0$ &  $0.4$ \\
Red Wine &  $1.0$  & $0.3$ \\
concrete & $1.0$  &  $0.3$ \\
yeast & $1.0$  &  $0.3$ \\
breast & $1.0$  &  $0.4$ \\
banknote & $1.0$  &  $0.2$ \\
\bottomrule
\end{tabular}
\end{table*}

\begin{table*}[!tbp]
\caption{Optimal parameters for REG-not-MIWAE.}
\label{table:params_optimal_reg_not_miwae}
\centering
\begin{tabular}{ccc}
\toprule
 & \multicolumn{2}{c}{REG-Not-MIWAE} \\\cmidrule{2-3}
Dataset & $\lambda$ & $\mathcal{P}$ \\\cmidrule{1-3}
White Wine  & $1.0$ &  $0.5$ \\
Red Wine &  $1.0$  & $0.6$ \\
concrete & $0.5$  &  $0.5$ \\
yeast & $1.0$  &  $0.5$ \\
breast & $1.0$  &  $0.6$ \\
banknote & $1.0$  &  $0.5$ \\
\bottomrule
\end{tabular}
\end{table*}

\enlargethispage*{500pt}
\clearpage

\section{Experimental results on partial VAEs}
\label{section:remaining_partial_experiments}
In this section we present additional exeprimental results for partial VAEs. Results on imputation quality, ELBO, and negative log-likelihood will be presented in Subsections ~\ref{section:additional_rmse}, ~\ref{sec:exp-additional}, and ~\ref{section:additional_nllh}, respectively. In Subsection ~\ref{sec:additional_ICs} additional Information Curves will be introduced. Experiment on the efficiency of acquired features will be presented in Subsection ~\ref{sec:additional_IC_experiment_efficiency}. Finally in Subsection ~\ref{sec:experiment_p_mechanisms} the imputation quality of different p-missingness mechanisms will be compared.

\subsection{Additional imputation quality Results}
\label{section:additional_rmse}
\begin{figure*}[!tb]
  \centering
  \setlength{\belowcaptionskip}{-0.2\baselineskip}
  \begin{subfigure}[c]{1.0\textwidth}
    \includegraphics[width=\textwidth]{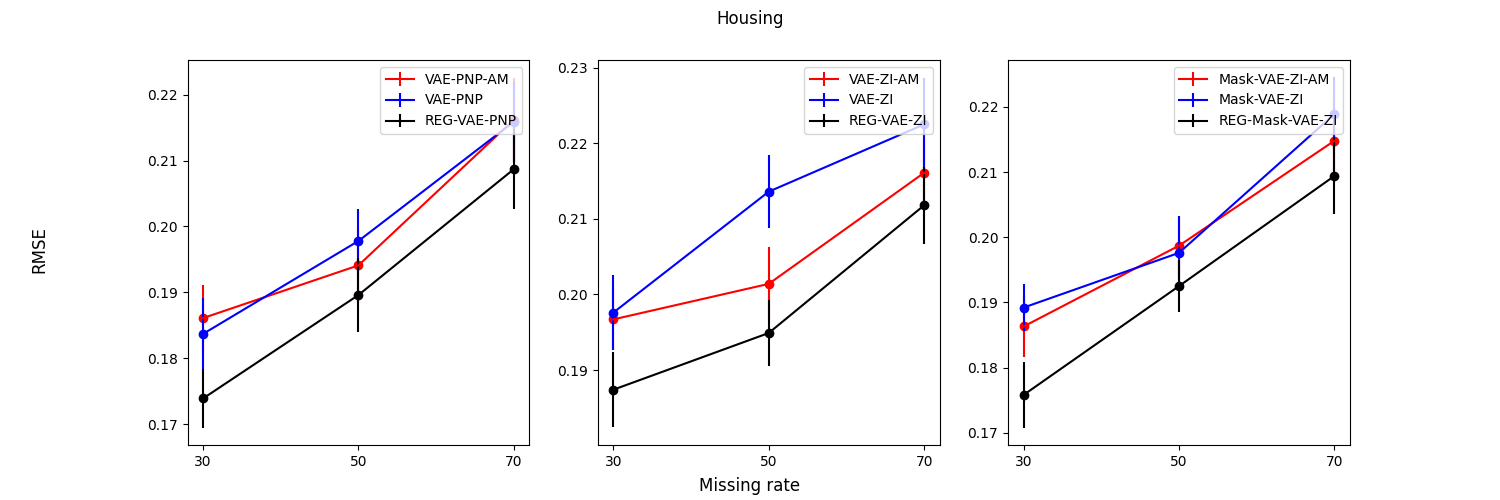}
  \end{subfigure} \\
  \begin{subfigure}[c]{1.0\textwidth}
    \includegraphics[width=\textwidth]{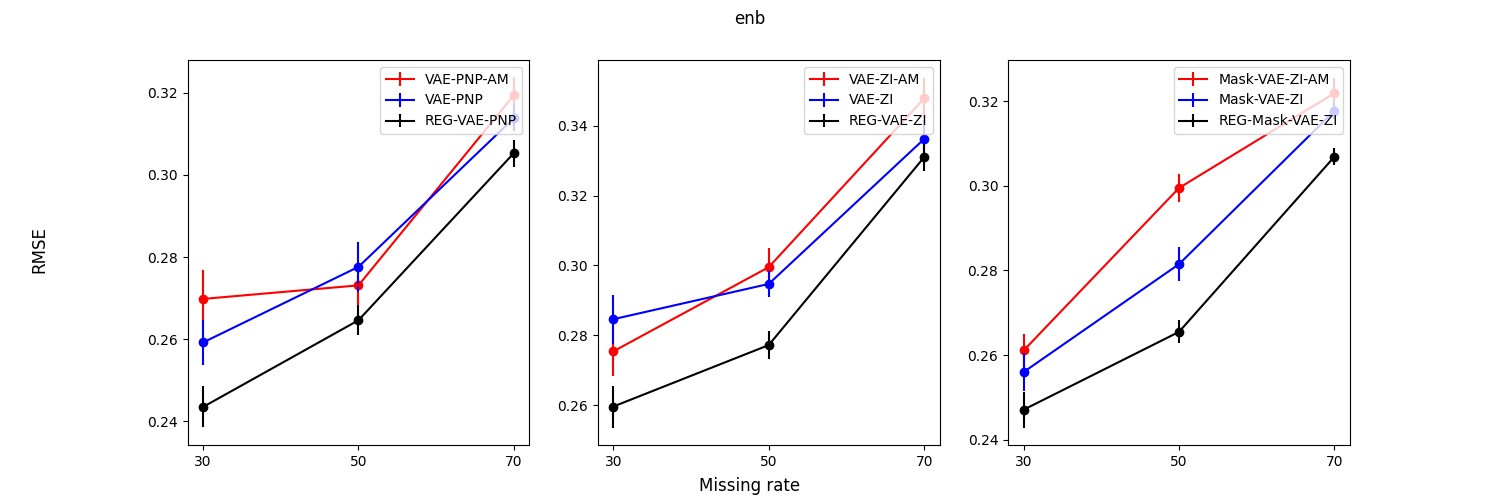}
  \end{subfigure}\\
  \begin{subfigure}[c]{1.0\textwidth}
    \includegraphics[width=\textwidth]{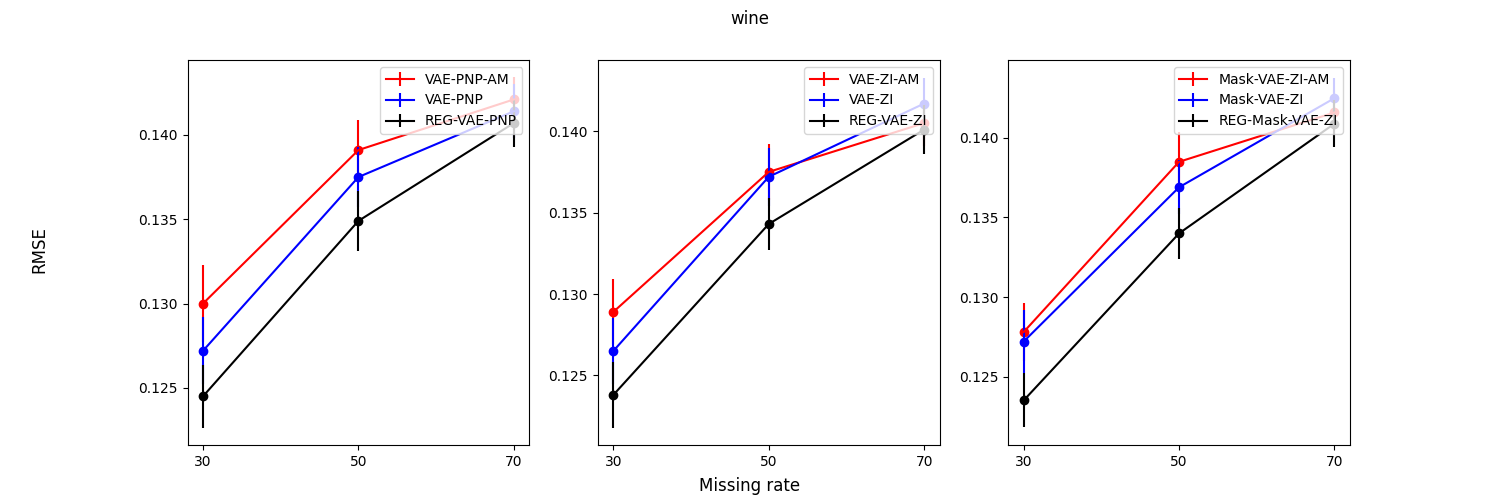}
  \end{subfigure}
  \caption{Imputation quality for various missing rates on different datasets.}
  \label{fig:graph_different_missing_rate}
  \vspace{-3mm}
\end{figure*}

To gain a better understanding of how different models perform in terms of imputation quality, refer to Figure ~\ref{fig:graph_different_missing_rate} for a visual comparison.

To compare the imputation quality of different models on data with 50 and 70 percent missingness, refer to Tables ~\ref{table:general_comparison_impute_qual_50} and ~\ref{table:general_comparison_impute_qual_70}, respectively.
\begin{table*}[!tbp]
\caption{Imputation quality (RMSE) for $50\%$ missingness on the test set. Smaller is better.}
\label{table:general_comparison_impute_qual_50}
\centering
\begin{tabular}{cccc}
\toprule
  & \multicolumn{3}{c}{Dataset} \\\cmidrule{2-4}
Model & Housing & Wine & enb \\\midrule
VAE-ZI-AM & $0.2014 \pm 0.0049$ & $0.1375 \pm 0.0017$ & $0.2995 \pm 0.0054$ \\
VAE-ZI & $ 0.2136 \pm 0.0048$ & $0.1372 \pm 0.0018$ & $ 0.2947 \pm 0.0037$ \\
REG-VAE-ZI & $\mathbf{0.1949\pm 0.0044}$ & $\mathbf{0.1343 \pm 0.0016}$ & $\mathbf{0.2772 \pm  0.0041 }$ \\[1mm]
Mask-VAE-ZI-AM & $0.1987 \pm 0.0046 $ & $0.1385 \pm 0.0019$ & $0.2995 \pm  0.0032$ \\
Mask-VAE-ZI & $ 0.1976 \pm 0.0056$ & $0.1369 \pm 0.0017$ & $0.2815 \pm  0.004$ \\
REG-Mask-VAE-ZI & $\mathbf{0.1925\pm 0.004}$ & $\mathbf{0.1340 \pm 0.0016 }$ & $\mathbf{0.2655 \pm 0.0027 }$ \\[1mm]
VAE-PNP-AM & $0.1941 \pm 0.0057$ & $0.1391 \pm 0.0018$ & $0.2731 \pm  0.0048 $ \\
VAE-PNP & $0.1978\pm 0.0048$ & $0.1375 \pm 0.0018$ & $0.2776 \pm 0.006$ \\
REG-VAE-PNP    & $\mathbf{0.1896 \pm 0.0056}$   & $\mathbf{0.1349 \pm  0.0018 }$   & $\mathbf{ 0.2646\pm 0.0037 }$  \\
\bottomrule
\end{tabular}
\end{table*}

\begin{table*}[!tbp]
\caption{Imputation quality (RMSE) for $70\%$ missingness on the test set. Smaller is better.}
\label{table:general_comparison_impute_qual_70}
\centering
\begin{tabular}{cccccccccc}
\toprule
  & \multicolumn{3}{c}{Dataset} \\\cmidrule{2-4}
Model & Housing & Wine & enb \\\midrule
VAE-ZI-AM & $0.2161 \pm 0.0045$ & $0.1405 \pm 0.0014$ & $0.3478 \pm 0.006$ \\
VAE-ZI & $ 0.2225 \pm 0.0062 $ & $ 0.1417 \pm 0.0016 $ & $ 0.3362 \pm 0.0053 $ \\
REG-VAE-ZI & $\mathbf{0.2118 \pm 0.0051 }$ & $\mathbf{0.1401 \pm 0.0015 }$ & $\mathbf{0.3310 \pm 0.0039 }$ \\[1mm]
Mask-VAE-ZI-AM & $0.2148 \pm 0.006  $ & $0.1416 \pm  0.0014 $ & $0.3219 \pm 0.0037  $ \\
Mask-VAE-ZI & $ 0.2189 \pm 0.0056 $ & $ 0.1425 \pm 0.0013 $ & $ 0.3177 \pm 0.0019 $ \\
REG-Mask-VAE-ZI & $\mathbf{0.2094 \pm 0.0058 }$ & $\mathbf{0.1409 \pm 0.0015 }$ & $\mathbf{0.3069 \pm 0.0020 }$ \\[1mm]
VAE-PNP-AM & $0.2162 \pm 0.0064 $ & $0.1421 \pm 0.0013 $ & $0.3194 \pm 0.0043 $ \\
VAE-PNP & $0.2159 \pm 0.0061 $ & $0.1414 \pm 0.0016 $ & $0.3138 \pm 0.0032 $ \\
REG-VAE-PNP    & $\mathbf{0.2087\pm 0.0061 }$   & $\mathbf{0.1407 \pm 0.0014 }$   & $\mathbf{0.3053 \pm 0.0033 }$   \\
\bottomrule
\end{tabular}
\end{table*}

\subsection{ELBO Results}
\label{sec:exp-additional}
To compare the ELBO Results of different models on data with 30, 50, and 70 percent missingness, refer to Tables ~\ref{table:general_comparison_elbo}, ~\ref{table:general_comparison_elbo_50}, and ~\ref{table:general_comparison_elbo_70}, respectively.

\begin{table*}[!tbp]
\setlength{\tabcolsep}{4pt}
\caption{ELBO values for $30\%$ missingness on the test set. Higher is better.}
\label{table:general_comparison_elbo}
\centering
\begin{tabular}{cccc}
\toprule
 & \multicolumn{3}{c}{Dataset} \\\cmidrule{2-4}
 Model & Housing & Wine & enb \\\midrule
 VAE-ZI-AM & $0.5766 \pm 0.244$ & $2.0158 \pm 0.086 $ & $-1.850 \pm 0.16 $ \\
 VAE-ZI & $ 0.4347 \pm 0.301$ & $2.0888 \pm 0.077$ & $-1.826 \pm 0.152$  \\
 REG-VAE-ZI & $\mathbf{1.1863\pm 0.183}$ & $\mathbf{2.115 \pm 0.074}$ & $\mathbf{-1.484 \pm 0.124 }$ \\[1mm]
 Mask-VAE-ZI-AM & $1.1026 \pm 0.219 $ & $2.1111 \pm 0.078$ & $-1.0338 \pm  0.108$ \\
 Mask-VAE-ZI & $0.9554 \pm 0.204$ & $2.1459 \pm 0.074$ & $\mathbf{-0.7451 \pm  0.099}$ \\
 REG-Mask-VAE-ZI & $\mathbf{1.368\pm 0.176}$ & $\mathbf{2.1591\pm 0.069}$ & $-0.7768 \pm 0.132$ \\[1mm]
 VAE-PNP-AM  & $0.6723 \pm 0.221$ & $1.9991 \pm 0.077$ & $-1.8753 \pm 0.146 $ \\
 VAE-PNP & $0.8378 \pm 0.220$ & $2.1126 \pm 0.080$ & $\mathbf{-1.0911 \pm 0.162}$ \\
 REG-VAE-PNP  & $\mathbf{1.2259 \pm 0.145}$ & $\mathbf{2.1497 \pm 0.077 }$ & $-2.0847 \pm 0.195$ \\
\bottomrule
\end{tabular}
\end{table*}

\begin{table*}[!tbp]
\caption{ELBO values for $50\%$ missingness on the test set. Higher is better.}
\label{table:general_comparison_elbo_50}
\centering
\begin{tabular}{cccc}
\toprule
  & \multicolumn{3}{c}{Dataset} \\\cmidrule{2-4}
Model & Housing & Wine & enb \\\midrule
VAE-ZI-AM & $-3.7005 \pm  0.2213$ & $-1.7670 \pm 0.0718$ & $-5.0869 \pm 0.1867$ \\
VAE-ZI & $-4.4814 \pm  0.1760$ & $ - 1.7888 \pm 0.0761$ & $-5.3655 \pm 0.2238$ \\
REG-VAE-ZI & $\mathbf{-3.5877\pm 0.1972}$ & $\mathbf{-1.7315 \pm 0.0664}$ & $\mathbf{-4.7483 \pm 0.1584 }$ \\[1mm]
Mask-VAE-ZI-AM & $-3.2981 \pm 0.2116 $ & $-1.7523 \pm 0.0759$ & $-3.9446 \pm 0.1027 $ \\
Mask-VAE-ZI & $-3.6595 \pm 0.2195$ & $-1.7234 \pm 0.073$ & $-4.0036 \pm 0.1353$ \\
REG-Mask-VAE-ZI & $\mathbf{-3.2136 \pm 0.189}$ & $\mathbf{- 1.7117 \pm 0.0678 }$ & $\mathbf{-3.8170\pm 0.0961}$ \\[1mm]
VAE-PNP-AM & $-3.4318 \pm 0.1909$ & $-1.7842 \pm 0.0710$ & $-4.4671 \pm 0.1548$ \\
VAE-PNP & $- 3.5437 \pm 0.2382$ & $\mathbf{- 1.7461 \pm 0.0796} $ & $-4.4415 \pm 0.1739$ \\
REG-VAE-PNP    & $\mathbf{-3.3533 \pm 0.1983}$   & $-1.7921 \pm 0.0736$   & $\mathbf{-4.3559 \pm 0.1445}$  \\
\bottomrule
\end{tabular}
\end{table*}

\begin{table*}[!tbp]
\caption{ELBO values for $70\%$ missingness on the test set. Higher is better.}
\label{table:general_comparison_elbo_70}
\centering
\begin{tabular}{cccccccccc}
\toprule
  & \multicolumn{3}{c}{Dataset} \\\cmidrule{2-4}
Model & Housing & Wine & enb \\\midrule
VAE-ZI-AM & $-7.9106\pm  0.2204$ & $-5.7689 \pm 0.0744$ & $-7.3366 \pm 0.0938$ \\
VAE-ZI & $ -8.3054 \pm 0.2454 $ & $ -5.7749 \pm 0.0776 $ & $ -7.8564 \pm  0.1402 $ \\
REG-VAE-ZI & $\mathbf{-7.9105 \pm 0.1975 }$ & $\mathbf{-5.7655 \pm 0.0686 }$ & $\mathbf{-7.3003 \pm 0.1123 }$ \\[1mm]
Mask-VAE-ZI-AM & $- 7.5994 \pm 0.1865  $ & $-5.7394 \pm 0.0800  $ & $-6.6569 \pm 0.1024  $ \\
Mask-VAE-ZI & $ -7.7758 \pm 0.2319 $ & $ -5.7450 \pm 0.0777 $ & $ -6.5186 \pm 0.0925 $ \\
REG-Mask-VAE-ZI & $\mathbf{- 7.5981 \pm 0.1906 }$ & $\mathbf{-5.7366 \pm 0.0725 }$ & $\mathbf{-6.4434\pm 0.0605 }$ \\[1mm]
VAE-PNP-AM & $- 7.6498\pm 0.2182$ & $-5.7437 \pm 0.0784 $ & $-6.8081 \pm 0.0832 $ \\
VAE-PNP & $-7.7355 \pm 0.1900 $ & $\mathbf{-5.7371 \pm 0.0793} $ & $-6.7737 \pm 0.0869 $ \\
REG-VAE-PNP    & $\mathbf{-7.7024 \pm 0.1907}$   & $-5.7432 \pm 0.0703$   & $\mathbf{-6.7537 \pm 0.0598 }$   \\
\bottomrule
\end{tabular}
\end{table*}

\subsection{Negative log-likelihood Results}
\label{section:additional_nllh}
To compare the negative log-likelihood Results of different models on data with 30, 50, and 70 percent missingness, refer to Tables ~\ref{table:general_comparison_negative_llh}, ~\ref{table:general_comparison_negative_llh_50}, and ~\ref{table:general_comparison_negative_llh_70}, respectively.

\begin{table*}[!tbp]
\caption{Negative log-likelihood of the imputation for $30\%$ missingness on the test set. Smaller is better.}
\label{table:general_comparison_negative_llh}
\centering
\begin{tabular}{cccc}
\toprule
  & \multicolumn{3}{c}{Dataset} \\\cmidrule{2-4}
Model & Housing & Wine & enb \\\midrule
VAE-ZI-AM & $8.7078 \pm 0.1623$ & $5.5788 \pm 0.0365$ & $9.0685 \pm 0.2709 $ \\
VAE-ZI & $8.7481 \pm 0.2155$ & $5.5336 \pm 0.0434$ & $9.4942 \pm 0.2704$ \\
REG-VAE-ZI & $\mathbf{8.2912\pm 0.1844}$ & $\mathbf{5.463 \pm 0.0315 }$ & $\mathbf{8.485 \pm 0.2118 }$ \\[1mm]
Mask-VAE-ZI-AM & $8.2416 \pm  0.1689 $ & $5.5555 \pm 0.0373$ & $8.5458 \pm  0.1923$ \\
Mask-VAE-ZI & $8.3882 \pm 0.1547$ & $5.5441 \pm 0.0379$ & $8.3209 \pm  0.1695$ \\
REG-Mask-VAE-ZI & $\mathbf{7.9411\pm 0.1583}$ & $\mathbf{5.4584 \pm 0.038 }$ & $\mathbf{7.9594 \pm 0.1691 }$ \\[1mm]
VAE-PNP-AM & $8.2228 \pm 0.1929 $ & $5.6153 \pm  0.0428$ & $8.8721 \pm 0.2657 $ \\
VAE-PNP & $8.1662 \pm 0.2123$ & $5.5330 \pm 0.0408$ & $ 8.4892 \pm 0.231$ \\
REG-VAE-PNP    & $\mathbf{7.8104 \pm 0.1575}$   & $\mathbf{5.4959 \pm 0.0351 }$   & $\mathbf{7.8604 \pm 0.1948 }$   \\
\bottomrule
\end{tabular}
\end{table*}

\begin{table*}[!tbp]
\caption{Negative log-likelihood of the imputation for $50\%$ missingness on the test set. Smaller is better.}
\label{table:general_comparison_negative_llh_50}
\centering
\begin{tabular}{cccc}
\toprule
  & \multicolumn{3}{c}{Dataset} \\\cmidrule{2-4}
Model & Housing & Wine & enb \\\midrule
VAE-ZI-AM & $6.3127 \pm 0.3459$ & $2.1951 \pm 0.0887$ & $10.7832 \pm 0.4077 $ \\
VAE-ZI & $7.1978 \pm 0.3527 $ & $2.1754 \pm 0.0928$ & $ 10.4297 \pm 0.2699$ \\
REG-VAE-ZI & $\mathbf{5.8644\pm 0.3053}$ & $\mathbf{2.0607 \pm 0.0807}$ & $\mathbf{9.1665 \pm 0.2834}$ \\[1mm]
Mask-VAE-ZI-AM & $6.1189 \pm 0.3379 $ & $2.2364 \pm 0.091$ & $9.3033 \pm 0.2193 $ \\
Mask-VAE-ZI & $6.0606 \pm 0.3861$ & $2.1687 \pm 0.0814$ & $9.4595 \pm 0.2962$ \\
REG-Mask-VAE-ZI & $\mathbf{5.6863\pm 0.2789}$ & $\mathbf{2.0505 \pm 0.0797 }$ & $\mathbf{8.3382 \pm 0.177}$ \\[1mm]
VAE-PNP-AM & $5.8178 \pm 0.4032$ & $2.2594 \pm 0.1042$ & $8.8578 \pm 0.3219$ \\
VAE-PNP & $6.0667 \pm 0.3288$ & $2.1951 \pm 0.0898$ & $9.2063 \pm 0.4398$ \\
REG-VAE-PNP    & $\mathbf{5.5196 \pm 0.3799}$  & $\mathbf{2.0809 \pm 0.0888 }$   & $\mathbf{8.2772 \pm 0.2446}$   \\
\bottomrule
\end{tabular}
\end{table*}

\begin{table*}[!tbp]
\caption{Negative log-likelihood of the imputation for $70\%$ missingness on the test set. Smaller is better.}
\label{table:general_comparison_negative_llh_70}
\centering
\begin{tabular}{cccc}
\toprule
  & \multicolumn{3}{c}{Dataset} \\\cmidrule{2-4}
Model & Housing & Wine & enb \\\midrule
VAE-ZI-AM & $5.2426 \pm 0.4443$ & $-1.3659 \pm 0.1074$ & $16.8757 \pm 0.7018$ \\
VAE-ZI & $ 5.9707 \pm 0.6453 $ & $ -1.2867 \pm 0.1194 $ & $ 15.4986 \pm 0.6149 $ \\
REG-VAE-ZI & $\mathbf{4.8169 \pm 0.4906 }$ & $\mathbf{-1.3841 \pm 0.0999 }$ & $\mathbf{14.8699 \pm 0.4466 }$ \\[1mm]
Mask-VAE-ZI-AM & $5.1422 \pm 0.5941  $ & $-1.2924 \pm 0.1050  $ & $13.8661 \pm 0.4227  $ \\
Mask-VAE-ZI & $ 5.5717 \pm 0.5782 $ & $ -1.2454 \pm 0.1034 $ & $ 13.3958 \pm 0.2104 $ \\
REG-Mask-VAE-ZI & $\mathbf{4.5924 \pm 0.5625}$ & $\mathbf{-1.3389 \pm  0.1048 }$ & $\mathbf{12.1905 \pm 0.2052 }$ \\[1mm]
VAE-PNP-AM & $5.3042 \pm 0.6447 $ & $-1.2684 \pm 0.1005 $ & $13.5855 \pm 0.4742 $ \\
VAE-PNP & $5.2626 \pm 0.6232 $ & $-1.3027 \pm 0.1128 $ & $12.9509 \pm 0.3531 $ \\
REG-VAE-PNP    & $\mathbf{ 4.523 \pm 0.6031 }$   & $\mathbf{-1.347 \pm 0.1051 }$   & $\mathbf{12.0404 \pm 0.3354 }$   \\
\bottomrule
\end{tabular}
\end{table*}

\subsection{Additional Information Curves}
\label{sec:additional_ICs}
\begin{figure*}[!tb]
  \centering
  \setlength{\belowcaptionskip}{-0.2\baselineskip}
  \begin{subfigure}[c]{0.32\textwidth}
    \includegraphics[width=\textwidth]{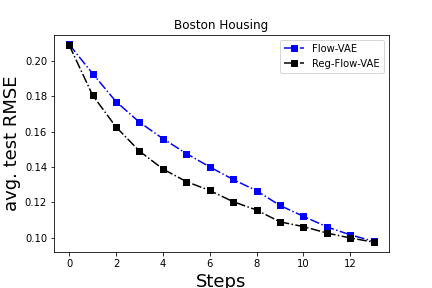}
    \subcaption{Boston housing}
  \end{subfigure}
  \begin{subfigure}[c]{0.32\textwidth}
    \includegraphics[width=\textwidth]{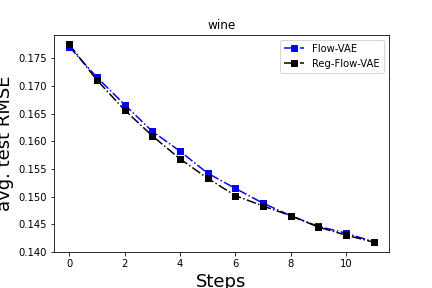}
    \subcaption{wine}
  \end{subfigure}
  \begin{subfigure}[c]{0.32\textwidth}
    \includegraphics[width=\textwidth]{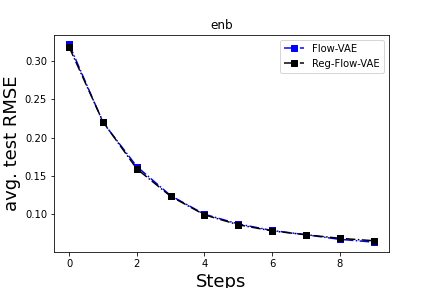}
    \subcaption{enb}
  \end{subfigure}
  \caption{Information curves on the test set for Flow-VAEs trained on the dataset with $30\%$ missingness.}
  \label{fig:add-ic-curves}
\end{figure*}

\begin{figure*}[htbp]
  \centering
  \begin{subfigure}[c]{0.3\textwidth}
    \includegraphics[width=\textwidth]{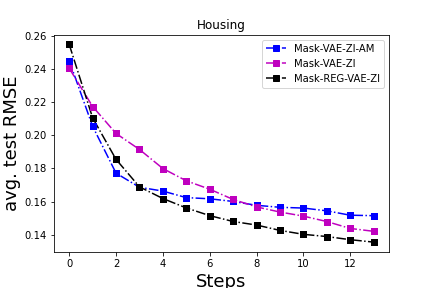}
    \subcaption{Housing}
  \end{subfigure}
  \begin{subfigure}[c]{0.3\textwidth}
    \includegraphics[width=\textwidth]{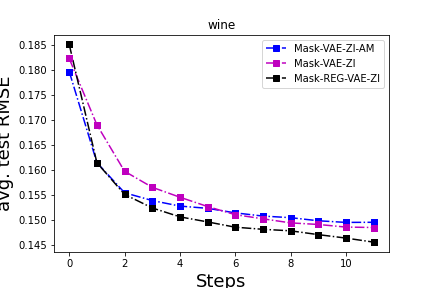}
    \subcaption{wine}
  \end{subfigure}
  \begin{subfigure}[c]{0.3\textwidth}
    \includegraphics[width=\textwidth]{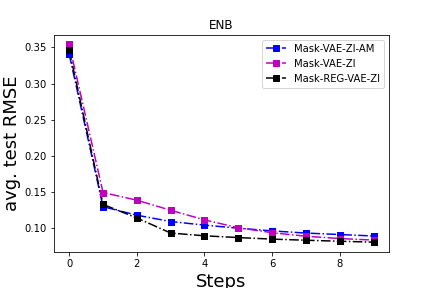}
    \subcaption{enb}
  \end{subfigure}

  \caption{Information curves on the test set for Mask-VAE-ZI, Mask-VAE-ZI-AM, and Mask-REG-VAE-ZI trained on the dataset with $30\%$ missingness.}
  \label{fig:mask_zi_comparison}
\end{figure*}

\begin{figure*}[htbp]
  \centering
  \begin{subfigure}[c]{0.3\textwidth}
    \includegraphics[width=\textwidth]{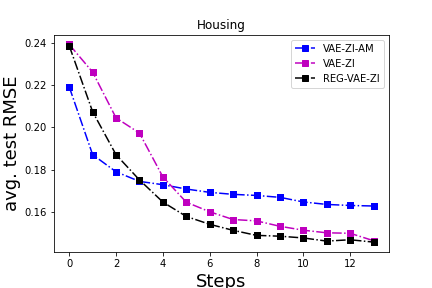}
    \subcaption{Housing}
  \end{subfigure}
  \begin{subfigure}[c]{0.3\textwidth}
    \includegraphics[width=\textwidth]{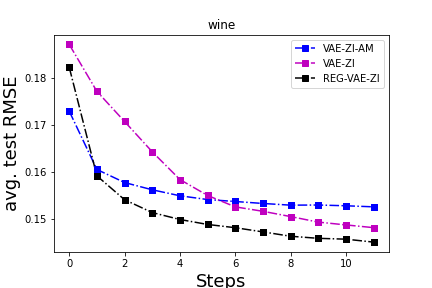}
    \subcaption{wine}
  \end{subfigure}
  \begin{subfigure}[c]{0.3\textwidth}
    \includegraphics[width=\textwidth]{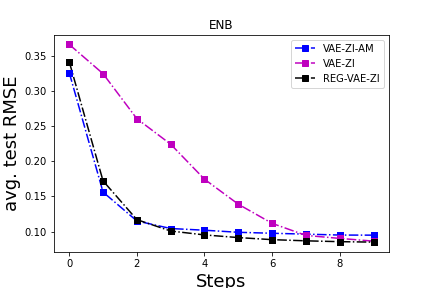}
    \subcaption{enb}
  \end{subfigure}
  \caption{Information curves on the test set for VAE-ZI, VAE-ZI-AM, and REG-VAE-ZI trained on the dataset with $30\%$ missingness.}
  \label{fig:zi_comparison}
\end{figure*}

\begin{figure*}[htbp]
  \centering
  \begin{subfigure}[c]{0.3\textwidth}
    \includegraphics[width=\textwidth]{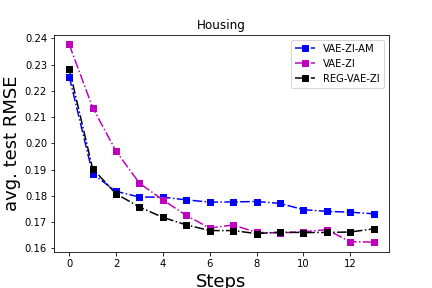}
    \subcaption{Housing}
  \end{subfigure}
  \begin{subfigure}[c]{0.3\textwidth}
    \includegraphics[width=\textwidth]{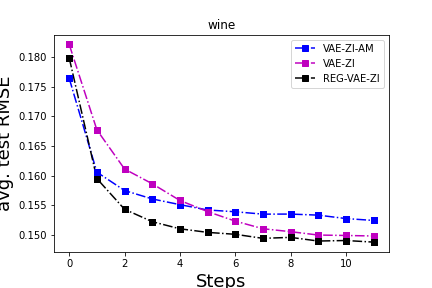}
    \subcaption{wine}
  \end{subfigure}
  \begin{subfigure}[c]{0.3\textwidth}
    \includegraphics[width=\textwidth]{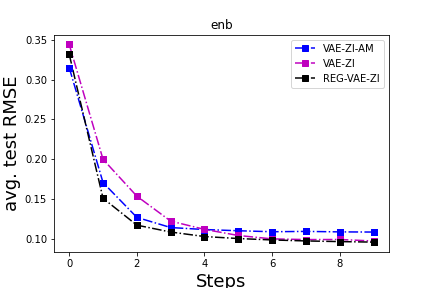}
    \subcaption{enb}
  \end{subfigure}
  \caption{Information curves on the test set for VAE-ZI, VAE-ZI-AM, and REG-VAE-ZI trained on the dataset with $50\%$ missingness.}
  \label{fig:zi_50_comparison}
\end{figure*}

\begin{figure*}[htbp]
  \centering
  \begin{subfigure}[c]{0.3\textwidth}
    \includegraphics[width=\textwidth]{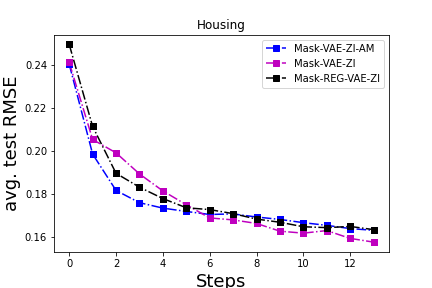}
    \subcaption{Housing}
  \end{subfigure}
  \begin{subfigure}[c]{0.3\textwidth}
    \includegraphics[width=\textwidth]{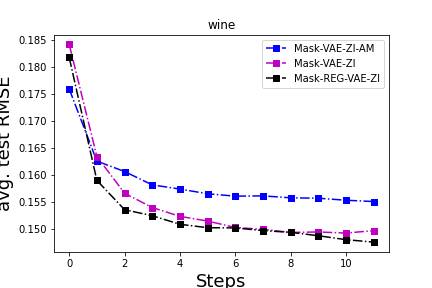}
    \subcaption{wine}
  \end{subfigure}
  \begin{subfigure}[c]{0.3\textwidth}
    \includegraphics[width=\textwidth]{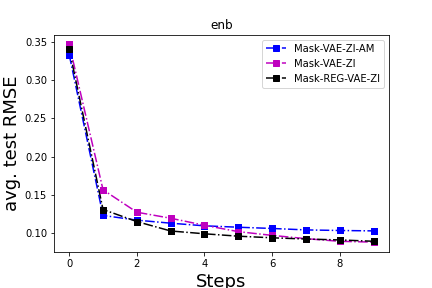}
    \subcaption{enb}
  \end{subfigure}
  \caption{Information curves on the test set for Mask-VAE-ZI, Mask-VAE-ZI-AM, and Mask-REG-VAE-ZI trained on the dataset with $50\%$ missingness.}
  \label{fig:mask_zi_50_comparison}
\end{figure*}

\begin{figure*}[htbp]
  \centering
  \begin{subfigure}[c]{0.3\textwidth}
    \includegraphics[width=\textwidth]{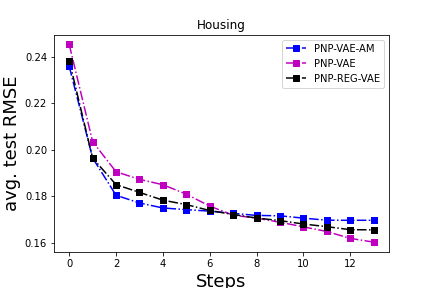}
    \subcaption{Housing}
  \end{subfigure}
  \begin{subfigure}[c]{0.3\textwidth}
    \includegraphics[width=\textwidth]{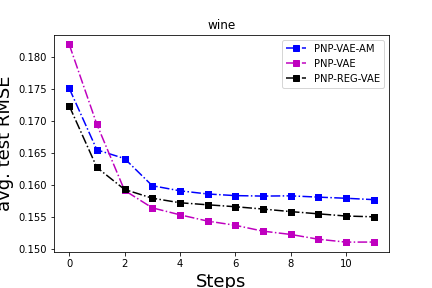}
    \subcaption{wine}
  \end{subfigure}
  \begin{subfigure}[c]{0.3\textwidth}
    \includegraphics[width=\textwidth]{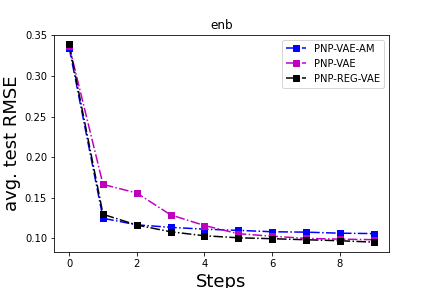}
    \subcaption{enb}
  \end{subfigure}
  \caption{Information curves on the test set for VAE-PNP, VAE-PNP-AM, and REG-VAE-PNP trained on the dataset with $50\%$ missingness.}
  \label{fig:pnp_50_comparison}
\end{figure*}

\begin{figure*}[htbp]
  \centering
  \begin{subfigure}[c]{0.3\textwidth}
    \includegraphics[width=\textwidth]{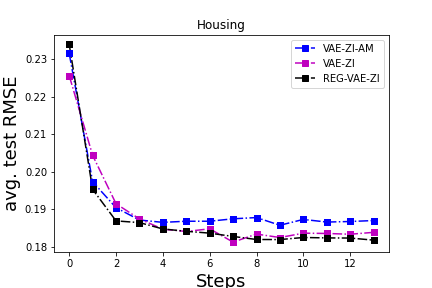}
    \subcaption{Housing}
  \end{subfigure}
  \begin{subfigure}[c]{0.3\textwidth}
    \includegraphics[width=\textwidth]{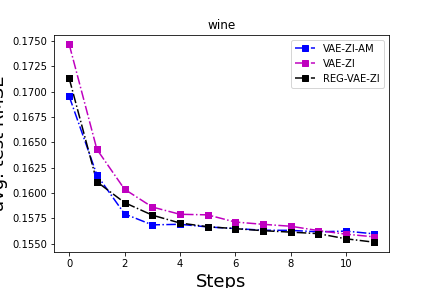}
    \subcaption{wine}
  \end{subfigure}
  \begin{subfigure}[c]{0.3\textwidth}
    \includegraphics[width=\textwidth]{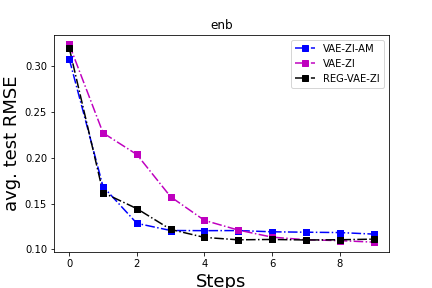}
    \subcaption{enb}
  \end{subfigure}
  \caption{Information curves on the test set for VAE-ZI, VAE-ZI-AM, and REG-VAE-ZI trained on the dataset with $70\%$ missingness.}
  \label{fig:zi_70_comparison}
\end{figure*}

\begin{figure*}[htbp]
  \centering
  \begin{subfigure}[c]{0.3\textwidth}
    \includegraphics[width=\textwidth]{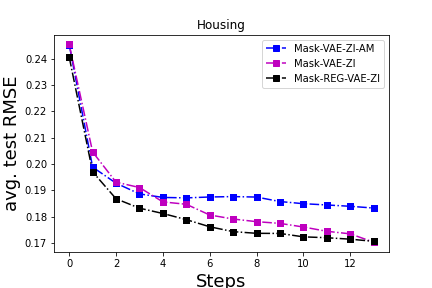}
    \subcaption{Housing}
  \end{subfigure}
  \begin{subfigure}[c]{0.3\textwidth}
    \includegraphics[width=\textwidth]{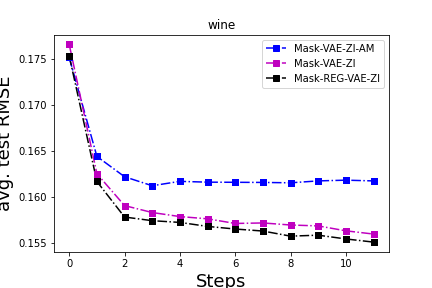}
    \subcaption{wine}
  \end{subfigure}
  \begin{subfigure}[c]{0.3\textwidth}
    \includegraphics[width=\textwidth]{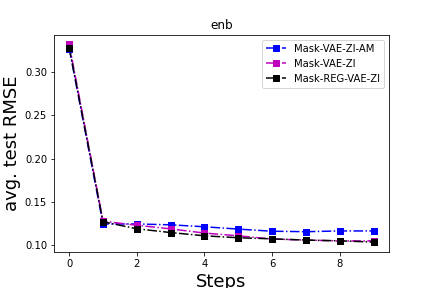}
    \subcaption{enb}
  \end{subfigure}
  \caption{Information curves on the test set for Mask-VAE-ZI, Mask-VAE-ZI-AM, and Mask-REG-VAE-ZI trained on the dataset with $70\%$ missingness.}
  \label{fig:mask_zi_70_comparison}
\end{figure*}

\begin{figure*}[htbp]
  \centering
  \begin{subfigure}[c]{0.3\textwidth}
    \includegraphics[width=\textwidth]{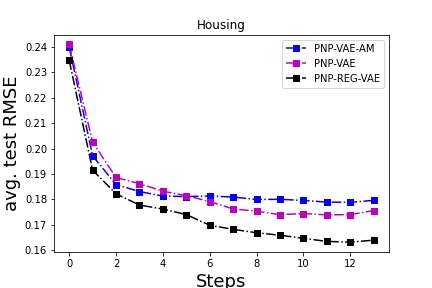}
    \subcaption{Housing}
  \end{subfigure}
  \begin{subfigure}[c]{0.3\textwidth}
    \includegraphics[width=\textwidth]{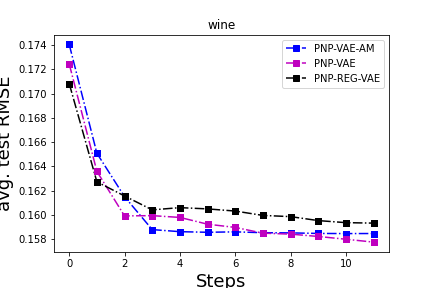}
    \subcaption{wine}
  \end{subfigure}
  \begin{subfigure}[c]{0.3\textwidth}
    \includegraphics[width=\textwidth]{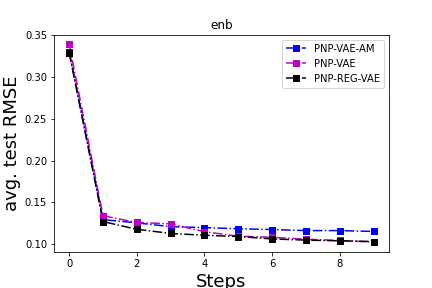}
    \subcaption{enb}
  \end{subfigure}
  \caption{Information curves on the test set for VAE-PNP, VAE-PNP-AM, and REG-VAE-PNP trained on the dataset with $70\%$ missingness.}
  \label{fig:pnp_70_comparison}
\end{figure*}

For Information Curves on data with 30 percent missingness see Figures ~\ref{fig:add-ic-curves}, ~\ref{fig:mask_zi_comparison} and ~\ref{fig:zi_comparison}.\\
For Information Curves on data with 50 percent missingness see Figures ~\ref{fig:pnp_50_comparison}, ~\ref{fig:mask_zi_50_comparison} and ~\ref{fig:zi_50_comparison}.\\
For Information Curves on data with 70 percent missingness see Figures ~\ref{fig:pnp_70_comparison}, ~\ref{fig:mask_zi_70_comparison} and ~\ref{fig:zi_70_comparison}.

\enlargethispage*{500pt}
\clearpage

\subsection{Experiment on the efficiency of the acquired features.}
\label{sec:additional_IC_experiment_efficiency}

\begin{figure*}[htbp]
  \centering
  \begin{subfigure}[c]{0.3\textwidth}
    \includegraphics[width=\textwidth]{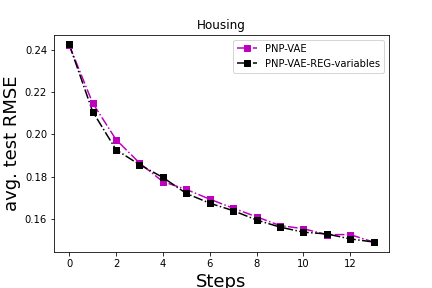}
    \subcaption{Housing}
  \end{subfigure}
  \begin{subfigure}[c]{0.3\textwidth}
    \includegraphics[width=\textwidth]{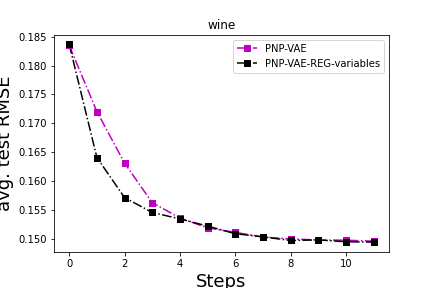}
    \subcaption{wine}
  \end{subfigure}
  \begin{subfigure}[c]{0.3\textwidth}
    \includegraphics[width=\textwidth]{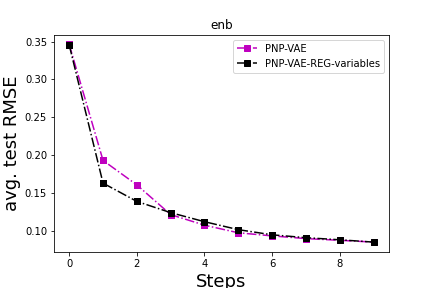}
    \subcaption{enb}
  \end{subfigure}
  \caption{Information curves of active variable selection on test data for VAE-PNP with features from REG-VAE-PNP.}
  \label{fig:pnp_comparison_var_reg}
\end{figure*}

\begin{figure*}[htbp]
  \centering
  \begin{subfigure}[c]{0.3\textwidth}
    \includegraphics[width=\textwidth]{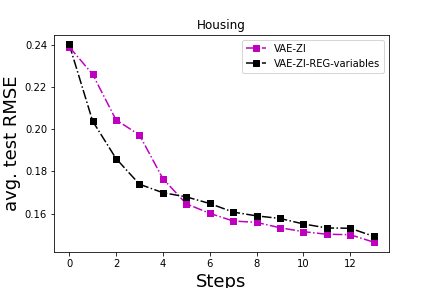}
    \subcaption{Housing}
  \end{subfigure}
  \begin{subfigure}[c]{0.3\textwidth}
    \includegraphics[width=\textwidth]{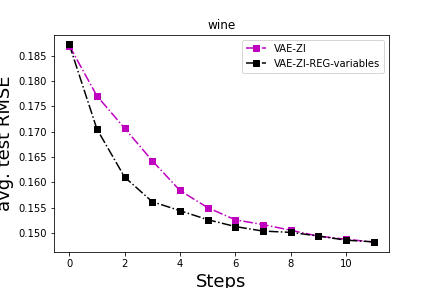}
    \subcaption{wine}
  \end{subfigure}
  \begin{subfigure}[c]{0.3\textwidth}
    \includegraphics[width=\textwidth]{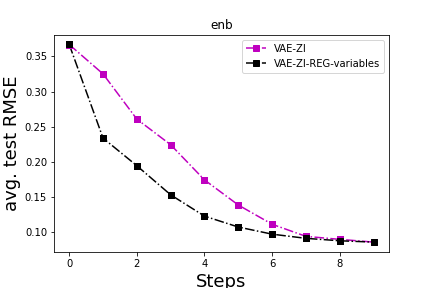}
    \subcaption{enb}
  \end{subfigure}
  \caption{Information curves of active variable selection on test data for VAE-ZI with features from REG-VAE-ZI.}
  \label{fig:zi_comparison_var_reg}
\end{figure*}

\begin{figure*}[htbp]
  \centering
  \begin{subfigure}[c]{0.3\textwidth}
    \includegraphics[width=\textwidth]{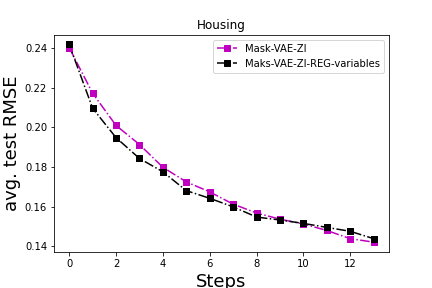}
    \subcaption{Housing}
  \end{subfigure}
  \begin{subfigure}[c]{0.3\textwidth}
    \includegraphics[width=\textwidth]{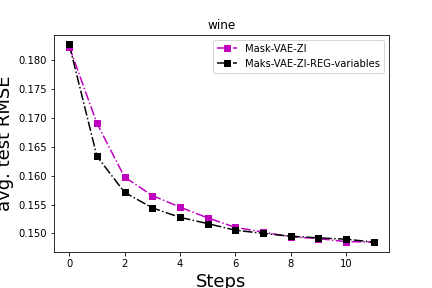}
    \subcaption{wine}
  \end{subfigure}
  \begin{subfigure}[c]{0.3\textwidth}
    \includegraphics[width=\textwidth]{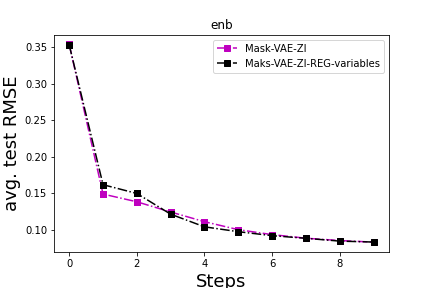}
    \subcaption{enb}
  \end{subfigure}
  \caption{Information curves of active variable selection on test data for Mask-VAE-ZI with features from Mask-REG-VAE-ZI.}
  \label{fig:mask_zi_comparison_var_reg}
\end{figure*}

To validate the hypothesis that posterior consistency regularization produces better feature selection for baselines, we conducted an experiment. During the active variable selection process, we input the acquired features from the regularized VAE-based imputation models into the baselines at each step of the process and evaluated the resulting change in the test RMSE score.\\
As shown in Figures \ref{fig:pnp_comparison_var_reg}, \ref{fig:zi_comparison_var_reg}, and \ref{fig:mask_zi_comparison_var_reg}, consistency regularization can provide more efficient variable selection for some classes of VAE-based imputation models, resulting in better RMSE scores.

\enlargethispage*{500pt}
\clearpage
\subsection{Experiment on different mechanisms of artificial missingness.}
\label{sec:experiment_p_mechanisms}
This experiment aims to compare the different mechanisms of artificial missingness for creating a subset of features $P$ from $Q$. The mechanisms tested were the \emph{half-feature mean}, \emph{all-feature mean}, \emph{half-feature variance}, \emph{all-feature variance}, and \emph{uniformly at random}. The \emph{half-feature mean} and \emph{half-feature variance} mechanisms are based on the mean or variance of the first half of the features, respectively. If the value of a particular feature in the first half is higher than its respective mean or variance, it is considered missing. The \emph{all-feature mean} and \emph{all-feature variance} mechanisms are constructed in the same manner, but based on the mean or variance of all features. The \emph{uniformly at random} mechanism corresponds to the artificial missingness described in line \ref{alg:missingness} of Algorithm \ref{alg:cap} and was used in previous experiments as the primary method of artificial missingness. For comparison, two baselines were included for each case of the VAE-based imputation model: \emph{no regularization} and \emph{no regularization with AM}. In this experiment, the true $30\%$ missingness of the dataset corresponds to the missingness created \emph{uniformly at random}.

The results, as shown in Tables \ref{table:general_comparison_impute_qual_30_different_mechanisms_boston}, \ref{table:general_comparison_impute_qual_30_different_mechanisms_enb}, and \ref{table:general_comparison_impute_qual_30_different_mechanisms_wine}, illustrate that using the artificial missingness based on the mean or variance can be beneficial in certain scenarios, particularly for more sophisticated models such as VAE-PNP and Mask-VAE-ZI, where information about missing features is utilized more effectively. However, this can also be advantageous in the case of VAE-ZI on some datasets. It is evident that the \emph{uniformly at random} mechanism consistently outperforms all other mechanisms for all VAE-based imputation models and datasets, where the true missingness mechanism is \emph{uniformly at random} across each dataset.

Note that the posterior relation~\eqref{eq:mcar_kl_equation} from Observation \ref{obs:posterior} may be applied to the subset $P$ created with MNAR missingness mechanisms based on the mean or variance values.

\begin{table*}[!tbp]
\caption{Imputation quality (RMSE) of VAE-based imputation models with different artificial missingness mechanisms on the test set of Housing dataset with $30\%$ missingness. Smaller is better.}
\label{table:general_comparison_impute_qual_30_different_mechanisms_boston}
\centering
\begin{tabular}{cccc}
\toprule
Type of artificial missingness &  REG-VAE-ZI & REG-Mask-VAE-ZI & REG-VAE-PNP   \\\cmidrule{1-4}
half-feature mean  &  $0.1996 \pm 0.0051$ &  $ 0.1882 \pm 0.0051 $ & $0.1835 \pm 0.0036 $   \\
all-feature mean &  $ 0.2 \pm 0.0059$ &  $ 0.1843 \pm 0.0048 $ & $0.181785 \pm 0.0043$   \\
half-feature variance &  $0.2012 \pm 0.0053$ &  $ 0.1886 \pm 0.0049 $ & $0.1849 \pm 0.005$  \\
all-feature variance &  $0.2030 \pm  0.0054 $ &  $ 0.1865 \pm 0.0053 $ & $0.1834 \pm 0.005$ \\
uniformly at random &  $0.1874  \pm 0.005$ &  $0.1758  \pm 0.005$ & $0.1739 \pm  0.0044$ \\
no regularization with AM &  $ 0.1967 \pm 0.004 $ &  $ 0.1863  \pm 0.0047 $ & $0.1861 \pm  0.005$ \\
no regularization &  $  0.1976 \pm 0.005 $ &  $ 0.1892  \pm 0.0036 $ & $0.1837 \pm 0.0055$ \\
\bottomrule
\end{tabular}
\end{table*}

\begin{table*}[!tbp]
\caption{Imputation quality (RMSE) of VAE-based imputation models with different artificial missingness mechanisms on the test set of ENB dataset with $30\%$ missingness. Smaller is better.}
\label{table:general_comparison_impute_qual_30_different_mechanisms_enb}
\centering
\begin{tabular}{cccc}
\toprule
Type of artificial missingness  &  REG-VAE-ZI & REG-Mask-VAE-ZI & REG-VAE-PNP   \\\cmidrule{1-4}
half-feature mean  &  $0.2809 \pm 0.0066$ &  $ 0.2538 \pm 0.003 $ & $0.2578 \pm 0.0039$   \\
all-feature mean &  $0.2696 \pm 0.0066$ &  $0.2530 \pm 0.0049 $ & $0.2485 \pm 0.0048$   \\
half-feature variance &  $0.2848 \pm 0.0059$ &  $ 0.2553 \pm 0.0027 $ & $0.263862 \pm 0.0046$  \\
all-feature variance &  $0.2830 \pm 0.0072$ &  $ 0.2571 \pm 0.0039 $ & $0.2576 \pm 0.0043$ \\
uniformly at random &  $0.2596  \pm 0.006$ &  $0.2471 \pm 0.0042$ & $0.2435 \pm 0.005$ \\
no regularization with AM &  $ 0.2754 \pm 0.007 $ &  $ 0.2611   \pm 0.004 $ & $0.2698  \pm  0.007$ \\
no regularization &  $  0.2846 \pm 0.007 $ &  $ 0.256  \pm 0.0045 $ & $0.2592 \pm  0.0055$ \\
\bottomrule
\end{tabular}
\end{table*}

\begin{table*}[!tbp]
\caption{Imputation quality (RMSE) of VAE-based imputation models with different artificial missingness mechanisms on the test set of Wine dataset with $30\%$ missingness. Smaller is better.}
\label{table:general_comparison_impute_qual_30_different_mechanisms_wine}
\centering
\begin{tabular}{cccc}
\toprule
Type of artificial missingness  &  REG-VAE-ZI & REG-Mask-VAE-ZI & REG-VAE-PNP   \\\cmidrule{1-4}
half-feature mean  &  $0.1264 \pm 0.0017$ &  $ 0.1267 \pm 0.0021 $ & $0.1273 \pm 0.002$   \\
all-feature mean &  $0.127 \pm 0.0021$ &  $ 0.1262 \pm 0.0018 $ & $0.1258 \pm 0.0019$   \\
half-feature variance &  $0.1267 \pm 0.0019$ &  $ 0.1256 \pm 0.0018 $ & $0.1257 \pm 0.002$  \\
all-feature variance &  $0.1272 \pm 0.002$ &  $ 0.1268 \pm 0.002 $ & $0.1268\pm 0.0019$ \\
uniformly at random &  $0.1238  \pm 0.002 $ &  $ 0.1235 \pm 0.0017 $ & $0.1245\pm 0.00185$ \\
no regularization with AM &  $ 0.1289 \pm 0.002 $ &  $ 0.1278   \pm 0.00185 $ & $0.13   \pm  0.0023$ \\
no regularization &  $  0.1265 \pm 0.002 $ &  $ 0.1272  \pm 0.002 $ & $0.1272 \pm  0.002 $ \\
\bottomrule
\end{tabular}
\end{table*}

\subsection{Imputation quality of the regularized models for different $\mathcal{P}$ parameters (probability of removing features)}
\label{sec:experiment_rmse_over_p_missing}

\enlargethispage*{1000pt}
\begin{figure*}[!tb]
  \centering
  \setlength{\belowcaptionskip}{-0.2\baselineskip}
  \begin{subfigure}[c]{1.0\textwidth}
    \includegraphics[width=\textwidth]{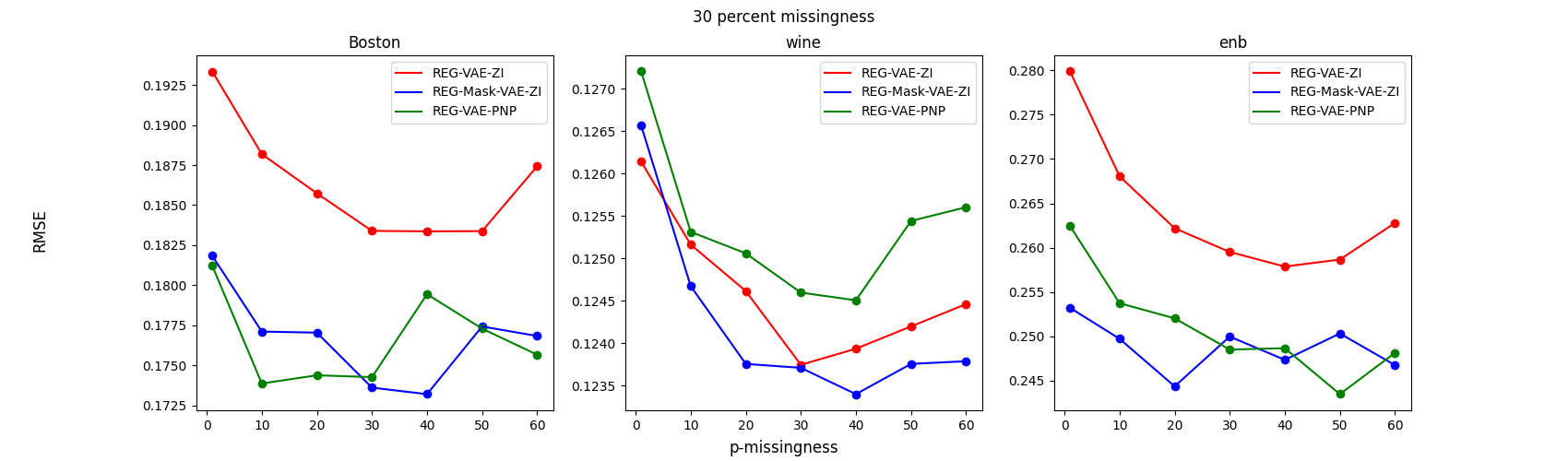}
  \end{subfigure} \\
  \begin{subfigure}[c]{1.0\textwidth}
    \includegraphics[width=\textwidth]{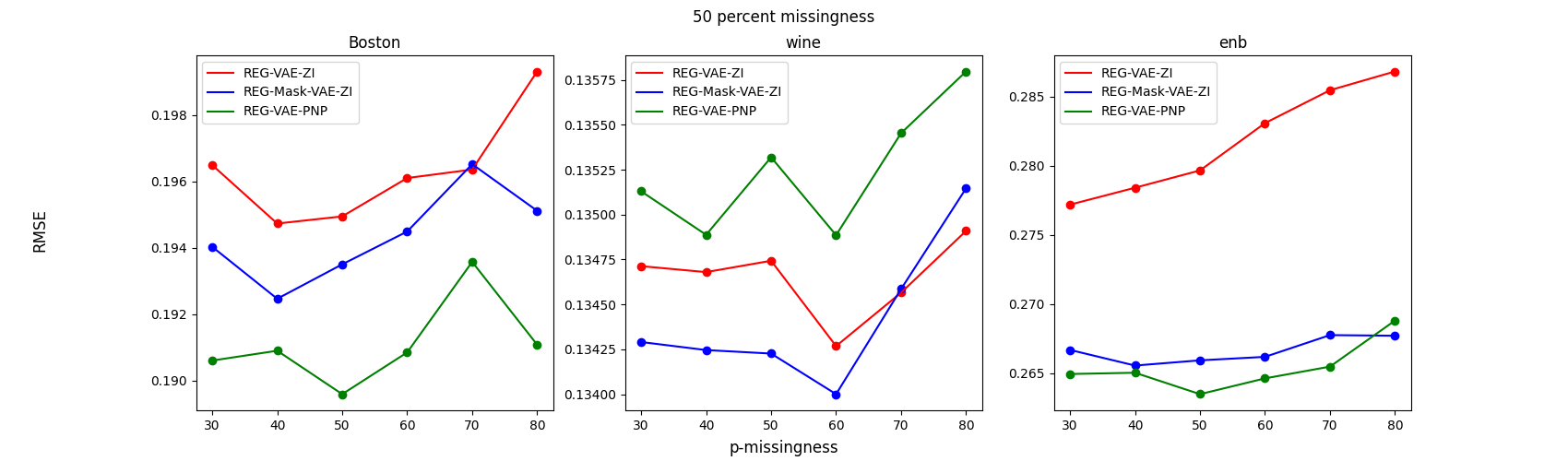}
  \end{subfigure}\\
  \begin{subfigure}[c]{1.0\textwidth}
    \includegraphics[width=\textwidth]{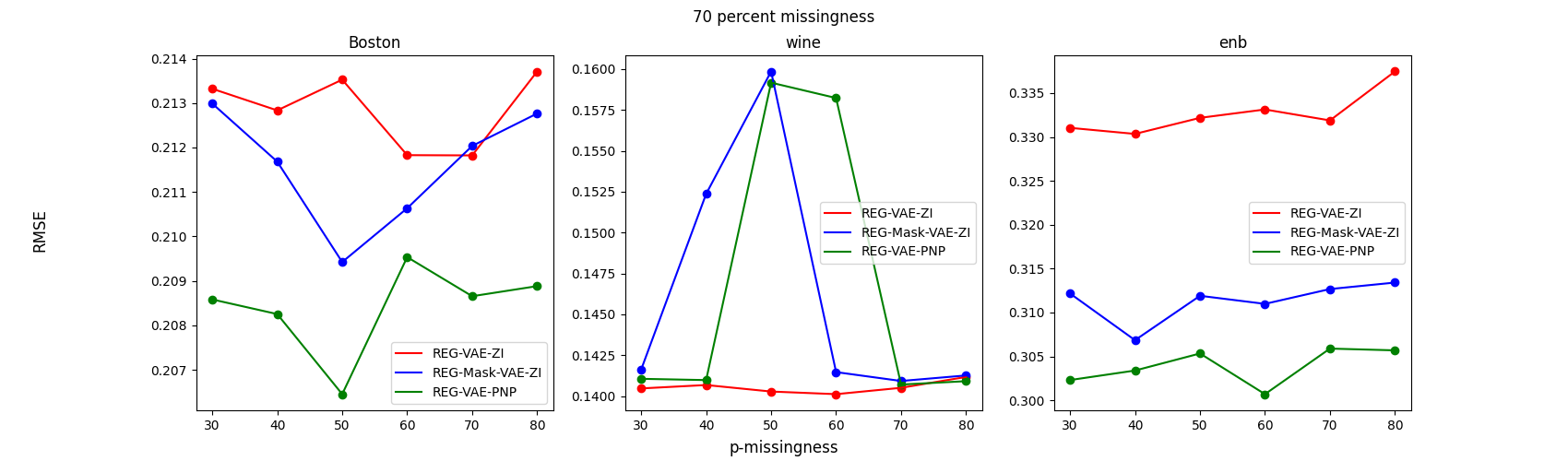}
  \end{subfigure}
  \caption{Imputation quality of the models over different $\mathcal{P}$ parameters for different percent missingness on the test data. p-missingness on the x-axis represents $\mathcal{P}$ parameter.}
  \label{fig:graph_different_p_missing_rmse}
  \vspace{-3mm}
\end{figure*}

The objective of this experiment was to evaluate how the performance of different regularized models is influenced by the $\mathcal{P}$ parameter, which characterizes the probability of removing the feature. The results presented in Figure~\ref{fig:graph_different_p_missing_rmse} show that the performance of the models varies significantly with $\mathcal{P}$ values, and the optimal $\mathcal{P}$ parameter falls within the range of the actual missingness plus or minus $20\%$.

\section{Flow-VAE in the MNAR setting}
\label{sec:flow_with_mnar}

We additionally considered the Flow-VAE model for one particular MNAR case, described in Section~\ref{sec:data}.
We can derive an ELBO for the joint likelihood of data $\vecx$ and missing mechanism $\textbf{m}$ as 
\begin{align}
\label{eq:mnar_elbo}
\log p_\theta(\vecx, \textbf{m}) &\geqslant \underbrace{\mathbb{E}_{q_\phi(\Z \mid \vecx, \vecm)}  \log p_\theta(\vecx \mid \vecz, \textbf{m})-\kl\left(q_\phi(\vecz \mid \vecx, \textbf{m}) \| p(\vecz)\right)}_{(A)}
& + \\
  &\phantom{\geqslant} \; \underbrace{\mathbb{E}_{q_\phi(\Z\mid \vecx, \vecm)} \log p_\gamma(\textbf{m} \mid \vecz)}_{(B)} \nonumber
\end{align}
Part $(A)$ from Equation~\eqref{eq:mnar_elbo} can be computed with the partial Flow-VAE and for computing part $(B)$ a Bernoulli distribution is used  (in the same way as in Not-MIWAE paper \cite{ipsen2021notMIWAE})
\begin{equation} \label{eq:bern_mask2}
p_\gamma(\textbf{m} \mid \vecx)=\operatorname{Bern}\left(\textbf{m} \mid \pi_\gamma(\vecx)\right)=\prod_{j=1}^D \pi_{\gamma, j}(\vecx)^{m_j}\left(1-\pi_{\gamma, j}(\vecx)\right)^{1-m_j},
\end{equation}
where $\pi_{\gamma, j}(\vecx)$ is calculated using self-masking and $D$ is a data dimension. \\
From the results in Table \ref{table:flow_mnar_comparison} we can observe that Flow-VAE outperforms Not-MIWAE for almost all datasets in the particular considered MNAR setting.

\begin{table*}[htbp]
\caption{Imputation quality (RMSE). Smaller is better.}
\label{table:flow_mnar_comparison}
\centering
\begin{tabular}{cccc}
\toprule
Dataset &  Not-MIWAE & VAE-flow-mnar \\\cmidrule{1-3}
Red Wine & $ 0.1594 \pm 0.0225 $ & $\mathbf{0.1174 \pm  0.0163 }$ \\
concrete & $ 0.2887 \pm  0.0387$ & $\mathbf{0.2126 \pm  0.0491 }$ \\
White Wine & $ 0.0891 \pm 0.01 $ & $\mathbf{0.0775 \pm  0.0066 }$ \\
banknote & $ 0.2459 \pm 0.0269 $ & $\mathbf{0.1651 \pm  0.0136 }$ \\
breast & $  0.1 \pm 0.0012 $ & $\mathbf{0.0689 \pm  0.0004 }$ \\
yeast & $  \mathbf{0.1363 \pm 0.001} $ & $0.1377 \pm  0.0009 $ \\
\bottomrule
\end{tabular}
\end{table*}

\section{Experimental results on full data}
\label{sec:experiments-full-data}
In addition, we explored the benefits of utilizing consistency regularization on fully observed data. Our analysis, presented in Tables \ref{table:full_comparison_impute_qual}, \ref{table:full_comparison_log_like}, and \ref{table:full_comparison_elbo}, indicates that considering the connection between $p_\parsDec(\vecz | \vecx_Q)$ and $p_\parsDec(\vecz | \vecx_P)$ marginally enhances the performance of the VAE model, even in a fully observed scenario. Notably, as the data is entirely observable in this case, $\vecx_Q$ is equivalent to $\vecx$, and $\vecx_P$ is generated using the standard approach.

\begin{table}[htbp]
\caption{Reconstruction quality (RMSE) on the test set. Smaller is better.}
\label{table:full_comparison_impute_qual}
\centering
\footnotesize
\begin{tabular}{ccc}
\toprule
Dataset &  VAE & REG-VAE  \\\cmidrule{1-3}
Housing  &   $0.0235 \pm 0.0004$ & $\mathbf{0.0232\pm 0.0002}$  \\
Wine &   $ 0.0281 \pm 0.0002 $ & $\mathbf{0.024 \pm 0.0002 }$ \\
enb &   $0.028 \pm 0.0003$ & $\mathbf{0.0271 \pm 0.0004 }$  \\
\bottomrule
\end{tabular}
\end{table}

\begin{table}[htbp]
\caption{Negative log-likelihood of the reconstruction on the test set. Smaller is better.}
\label{table:full_comparison_log_like}
\centering
\footnotesize
\begin{tabular}{ccc}
\toprule
Dataset &  VAE & REG-VAE  \\\cmidrule{1-3}
Housing  &   $-11.8111 \pm 0.0863$ & $\mathbf{-11.8768\pm 0.049}$  \\
Wine &   $-9.5999 \pm 0.0421$ & $\mathbf{-9.6085 \pm 0.0372 }$ \\
enb &  $-8.3871 \pm 0.0423$ & $\mathbf{-8.5163 \pm 0.0536 }$ \\
\bottomrule
\end{tabular}
\end{table}

\begin{table*}[htbp]
\caption{ELBO values on the test set. Higher is better.}
\label{table:full_comparison_elbo}
\centering
\begin{tabular}{ccc}
\toprule
Dataset & VAE & REG-VAE  \\\cmidrule{1-3}
Boston Housing  &    $ 8.4798 \pm 0.0814$ & $\mathbf{8.5286\pm 0.0885}$  \\
Wine &   $8.1021 \pm 0.0481$ & $\mathbf{8.1113 \pm 0.0497 }$ \\
enb &   $3.9105 \pm 0.0280$ &$\mathbf{3.9117 \pm 0.0169 }$  \\

\bottomrule
\end{tabular}
\end{table*}

\end{document}